\documentclass[journal]{IEEEtran}

\usepackage{glossaries}
\usepackage{graphicx}
\usepackage{caption}
\usepackage{subcaption}
\usepackage{amsfonts}
\usepackage{amsmath}
\usepackage{tabularx}

\DeclareMathOperator*{\argmin}{arg\,min}
\newcommand{\mb}[1]{\mathbf{#1}}

\newacronym{ivc}{IVC}{Inferior Vena Cava}
\newacronym{pv}{PV}{Pulmonary Vein}
\newacronym{dof}{DOF}{Degrees of Freedom}
\newacronym{afib}{Afib}{Atrial fibrillation}

\ifCLASSINFOpdf
\else
\fi
\begin{document}

\title{Magnetic Ball Chain Robots for Cardiac Arrhythmia Treatment}

\author{Giovanni~Pittiglio,~\IEEEmembership{Member,~IEEE,}
        Fabio~Leuenberger, Margherita~Mencattelli, Max~McCandless, Edward~O'Leary,
        and~Pierre~E.~Dupont,~\IEEEmembership{Fellow,~IEEE}
\thanks{G. Pittiglio was with the Department of Cardiovascular Surgery, Boston Children’s Hospital, Harvard Medical School. He is now with Department of Robotics Engineering, Worcester Polytechnic Institute (WPI), Worcester, MA 01609, USA. Email: {\tt \small gpittiglio@wpi.edu}}
\thanks{F. Leuenberger, M. McCandless, and P. E. Dupont are with the Department of Cardiovascular Surgery, Boston Children’s Hospital, Harvard Medical School, Boston, MA 02115, USA. Email: {\tt\small \{fabio.leuenberger, max.mccandless, pierre.dupont\}@childrens.harvard.edu}}
\thanks{M. Mencattelli was with the Department of Cardiovascular Surgery, Boston Children’s Hospital, Harvard Medical School. She is now with the Minimally Invasive Therapies Group - Surgical Innovation - Medtronic, Boston, MA 02210, USA : {\tt \small marghemenc@gmail.com}}
\thanks{E. O'Leary is with the Department of Cardiology, Boston Children’s Hospital, Harvard Medical School, Boston, MA 02115, USA. {\tt\small edward.oleary@cardio.chboston.org}}
\thanks{This work was supported by the National Institutes of Health under grant R01HL124020.}}


\maketitle

\begin{abstract}
This paper introduces a novel magnetic navigation system for cardiac ablation. The system is formed from two key elements: a magnetic ablation catheter consisting of a chain of spherical permanent magnets; and an actuation system comprised of two cart-mounted permanent magnets undergoing pure rotation. The catheter design enables a large magnetic content with the goal of minimizing the footprint of the actuation system for easier integration with the clinical workflow. We present a quasi-static model of the catheter, the design of the actuation units, and their control modalities. Experimental validation shows that we can use small rotating magnets (119mm diameter) to reach cardiac ablation targets while generating clinically-relevant forces. Catheter control using a joystick is compared with manual catheter control. While total task completion time is similar, smoother navigation is observed using the proposed robotic system. We also demonstrate that the ball chain can ablate heart tissue and generate lesions comparable to the current clinical ablation catheters. 
\end{abstract}


\IEEEpeerreviewmaketitle

\section{Introduction}
\IEEEPARstart{A}{trial} fibrillation is the most prevalent form of persistent cardiac arrhythmia with about 6 million cases each year in the USA \cite{Kornej2020}. When affected by this condition, the patient experiences uncoordinated heart beats, caused by faulty electrical conductivity of the myocardium (heart tissue) at the level of the atria. 
Cardiac ablation is a minimally-invasive procedure aiming at ablating the internal heart walls to restore healthy electrical conduction. This is associated with a lower recurrency rate compared to pharmacological treatments \cite{Prystowsky2015} and so ablation is commonly performed when medications fail.

\begin{figure}[t]
\begin{center}
\includegraphics[width=\columnwidth, bb=0 0 250 200]{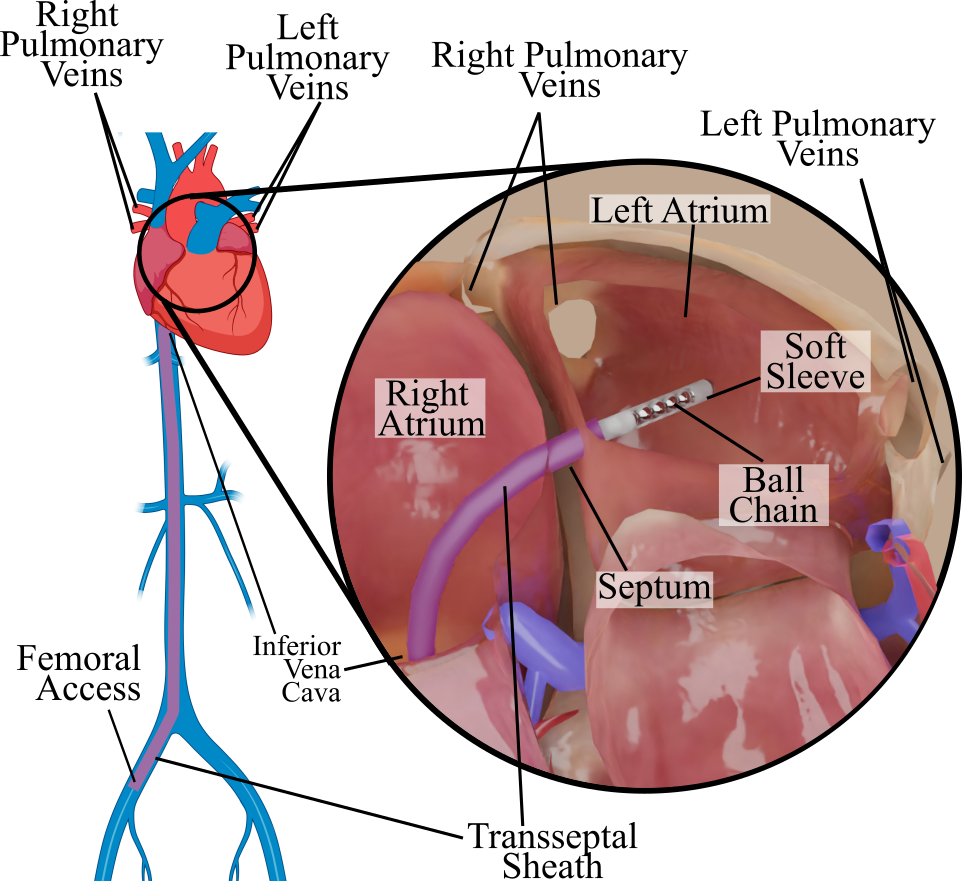}
\end{center}
\caption{Schematic representation of ball chain magnetic catheter introduced in the left atrium by femoral access.}
\label{fig:navigation_figure}
\end{figure}

Ablation procedures require a long flexible catheter with an ablating tip to reach the left atrium - where most arrhythmias originate. To do so, a sheath is inserted in the femoral vein (femoral access) and navigated to the right atrium (see Fig. \ref{fig:navigation_figure}). A small puncture is made in the atrial septum - the wall separating the left and right atria - and a sheath is extended slightly into the left atrium. Mapping and ablation catheters are inserted through the sheath to reach left atrium.

The clinician's task is to position the tip of the ablation catheter on the heart wall to create a set of connected lesions on the wall by ablating at contiguous spots. Electrical triggers, causing arrhythmic heart beating, mostly originate from the pulmonary veins \cite{Palama2022}. Therefore, a common procedure is to create lesions which enclose the pulmonary veins (left and right). This procedure is complex to master and both physically and mentally demanding; the clinician manipulates a tendon-actuated catheter from outside the body of the patient so as to steer the catheter's tip. Simultaneously, they need to carefully regulate the applied tip force to avoid burning through the wall of the heart. Clinical mapping systems provide a computer graphics image of the heart and catheter which the clinician uses for real-time localization and force sensing.

In the context of ablation procedures, robotics has the potential of producing more consistent ablations, reducing the operator's fatigue, extending the catheter's reach, and reducing the use of fluoroscopy \cite{Bassil2020}. Clinically-approved robotic systems for ablation include those of Hansen Medical \cite{Hlivak2011} and Stereotaxis \cite{Faddis2002}.

While no longer available, the Hansen Medical robotic system consisted of two telescoping tendon-actuated sheaths through which any standard ablation catheter could be deployed. A significant challenge of this approach is that the disposable steerable sheaths were costly and still required the hospital to buy an ablation catheter. 

In contrast, the Stereotaxis system replaces the standard ablation catheter with a relatively inexpensive design needing no tendon actuation, but instead incorporating permanent magnets embedded in its tip \cite{Faddis2002}. To steer the catheter tip, an external actuation system is used, comprising two robotic arms that move permanent magnets around the patient's body. This approach is also taken in other endoluminal procedures, e.g., in the digestive tract \cite{Barducci2019, Wright2015} and bronchi \cite{Pittiglio2023}. 

For ablation procedures, magnetic actuation has attracted particular attention \cite{Faddis2002, Chautems2017, Heemeyer2023, Piskarev2022} because magnetic forces and torques applied to the catheter tip can create more stable contact with the wall of the beating heart compared to forces exerted at the catheter's base. Since ablations involve remaining at a single spot on the heart wall for up to one minute, contact stability is very important. Moreover, tip pulling enables the use of softer catheters and reduces the risk of perforation \cite{Bassil2020}.

Despite the modest contact forces required for ablation (10-20g) \cite{Ariyarathna2018, Issa2019}, it has previously posed challenging to create a small external actuation system that can produce the desired workspace and tip forces. For example, the Stereotaxis system uses two robot arms carrying 800lb magnets \cite{Creighton2005}. The overall size of the system poses challenges in integration with the current clinical workflow, due size and weight.

The contribution of this paper is to propose and validate an alternative approach which avoids the high costs of additional steerable sheaths while also enabling the use of magnetic robotic steering with compact cart-mounted external magnets. The magnetic  ablation catheter is comprised of a magnetic ball chain \cite{Pittiglio2023d, ODonoghue2013} covered by a soft sleeve (see Fig. \ref{fig:navigation_figure}).  
Ball chains contain a higher magnetic content (75\% of the volume), compared to designs which use one or several cylindrical magnets \cite{Edelmann2017, Jeon2018} (10-20\%) or a magnetized polymeric matrix \cite{Kim2022, Pittiglio2022} (40\%). This design enables tighter bending \cite{Pittiglio2023d, Pittiglio2023c} and higher tip forces, which permits the use of smaller actuation systems. When compared to actuating magnetic polymers, we see an 87.5\% reduction in the required actuating field.

\begin{figure}[t]
\begin{center}
\includegraphics[width=\columnwidth, bb=0 0 250 200]{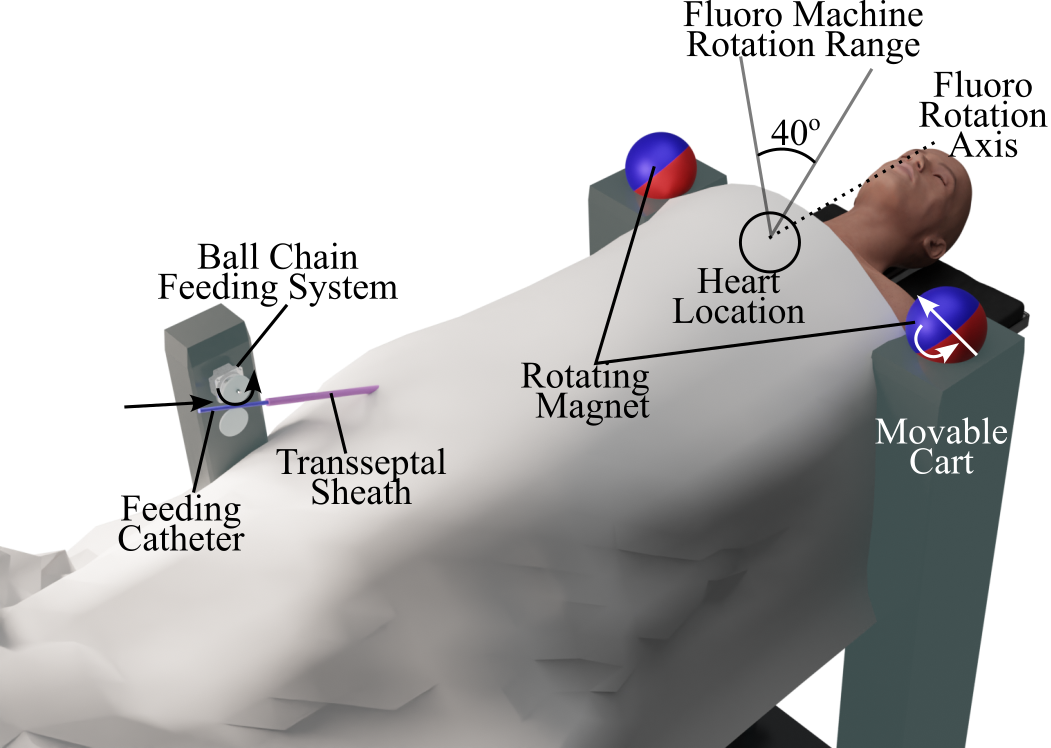}
\end{center}
\caption{Schematic representation of the magnetic navigation system for cardiac ablation.}
\label{fig:platform}
\end{figure}

To minimize the footprint of our actuation units, we hypothesized that appropriate actuation torques and forces could be produced using a pair of cart-mounted magnets undergoing pure rotation which are positioned to the left and right sides of the patient (Fig. \ref{fig:platform}). While the use of fluoroscopy is declining in ablation procedures, this configuration allows for imaging 40$^\circ$ about the longitudinal axis of the patient as shown in Fig. \ref{fig:platform}.



The remainder of the paper is organized as follows. Section \ref{sec:ball_chain} describes a quasi-static model for the ball chain, originally formulated in \cite{Pittiglio2023d}. Section \ref{sec:actuation} describes a scaled version of the actuation system which is used for experimental validation. In Section \ref{sec:control}, the implementation of teleoperated catheter control is described. 

Section \ref{sec:experiments} presents experimental validation. First, we measure the applied force on our target locations, using a force sensor. We use these measurements to extrapolate the dimensions of the full-scale actuation units for clinical applications (Section \ref{sub:force}). In Section \ref{sub:workspace}, we demonstrate that the proposed catheter design is capable of large deflections and that we can fabricate an elastic sleeve which does not reduce the chain's bending capability. Section \ref{sub:robotic_vs_manual} reports comparison of the robotic platform with a manual catheter. Finally, in Section \ref{sub:ablation}, we demonstrate that the proposed catheter is able to ablate heart tissue without requiring further on board electronics, thus, with no reduction of the catheter's bending capabilities. Conclusions are presented in the final section of the paper.

\section{Magnetic Ball Chains}
\label{sec:ball_chain}
\begin{figure}[t]
    \centering
    \includegraphics[width=0.75\columnwidth, bb=0 0 250 200]{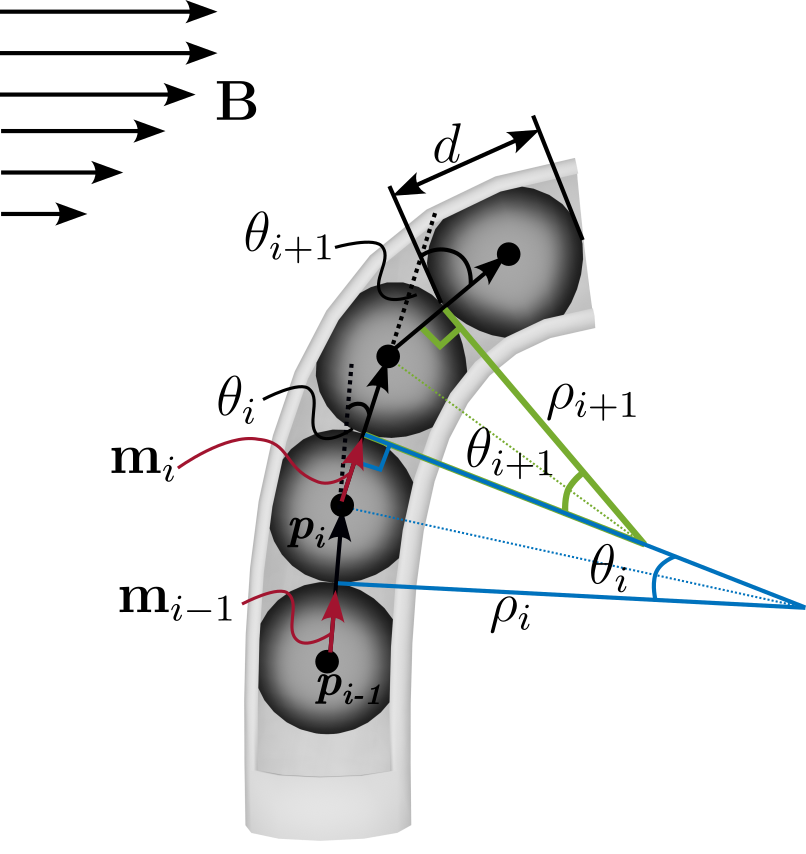}
    \caption{Description of magnetic ball chains. Applied field ($\mb B$) represented as a generic non-homogeneous vector field.}
    \label{fig:model_description}
\end{figure}

As shown in Fig. \ref{fig:model_description}, the spherical permanent magnets of a ball chain attract each other into a chain structure. When no external magnetic field is applied the chain forms a linear structure. When an external field is applied the induced torque leads the balls to align their magnetic dipole ($\mb{m}_i$) with the applied magnetic field ($\mb{B}$). 

The shape of the chain can be computed as the static equilibrium torque balance that minimizes the total energy of the system. The components of energy arise from the torques exerted by the spheres on each other, those generated on the spheres by the external field, the bending of the elastic sleeve encasing the chain of spheres and from any other externally applied forces or torques \cite{Pittiglio2023d}. 

In the following, we assume that the only external loading is that due to gravity. We also assume that there is no friction between the spheres and between the spheres and the elastic sleeve. All magnets involved are modelled as magnetic dipoles. In this work, a cylindrical magnet is used for actuation. A magnet of this shape can be modelled as a dipole at a distance larger than the diameter of its bounding sphere \cite{Petruska2013}, which is satisfied in our experiments. The balls in the chain can be accurately modelled as dipoles, since they are spherical \cite{Petruska2013}. 
It is assumed that there is no twisting in the elastic sleeve and that its curvature can be modeled as piecewise constant.

We define a body frame in each sphere in which the $\mb{z}$ axis is aligned with its magnetic dipole. In a world frame, the configuration of each sphere can be described by the position, $\mb{p}_i \in \mathbb{R}^3$ and orientation, $\mb R_i \in SO(3)$ of its body frame. 

We can write the potential energy of the ball chain, $U_b$, as the sum of energies from each pair of balls acting on each other:
\begin{eqnarray}
U_{b} &=& -\sum_{i = 1}^n \sum_{j = i + 1}^n {} \mb{m}_j \cdot \mb{B}(\mb{p}_j -  \mb{p}_i, \mb{m}_i)  \\
&=& -\sum_{i = 1}^n \sum_{j = i + 1}^n {} |\mb{m}_j|{} \left(\mb R_j \mb e_3 \right) \cdot \mb{B}(\mb{p}_j -  \mb{p}_i, |\mb{m}_i|{} \left(\mb R_i \mb e_3 \right) ) \nonumber
\end{eqnarray}
where the field of a magnet of magnetic dipole $\mb m$ at a distance $\mb r$ from its center is computed using the dipole model \cite{Petruska2013}
\begin{equation}
    \label{eq:dipole}
    \mb{B}(\mb{r}, \mb{m}) = \frac{|\mb{m}| \mu_0}{4 \pi |\mb{r}|^3} \left(3 \hat{\mb{r}} \hat{\mb{r}}^T - I \right) \hat{\mb{m}}
\end{equation}
and $\mb{e}_i \in \mathbb{R}^3$ is the $i$th element of the canonical basis of $\mathbb{R}^3$
Note that the energy is expressed with respect to the ball chain configuration parameters, $(\mb{p}_i, \mb R_i)$ by using the fact that $\mb{m}_i=|\mb{m}_i|\hat{\mb{m}}_i = |\mb{m}_i| \left(\mb R_i \mb e_3 \right)$. The symbol $\hat{\cdot}$ indicates the vector's direction and $|\cdot|$ its modulus.

Similarly, the potential energy between each ball and the external permanent magnet of magnetic dipole $\mb{m}_a$ in position $\mb{p}_a$ in world frame is
\begin{eqnarray}
U_{e} &=& -\sum_{i = 1}^n {} \mb{m}_i \cdot {} \mb{B}(\mb{p}_i - \mb{p}_a, \mb{m}_a )  \\
&=&-\sum_{i = 1}^n {} \mb{m}_i \cdot {} \mb{B}(\mb{p}_i - \mb{p}_a, |\mb{m}_a| \left(\mb R_a \mb e_3 \right) ) \nonumber
\end{eqnarray}

While we show that we can produce elastic sleeves of negligible stiffness (see Section \ref{sec:experiments}), we include its elastic energy due to bending here. We discretize the sleeve into arcs spanning the contact points between adjacent pairs of balls (Fig. \ref{fig:model_description}). The center line of the cylindrical sleeve section for ball $i$ is assumed to have a constant radius of curvature, $\rho_i$, and to span an angle, $\theta_i$. 

These variables can be expressed in terms of the shape parameters. The angle $\theta_i$ is defined by
\begin{equation}
\tan(\theta_i) = \frac{(\mb{p}_{i + 1}-\mb{p}_i) \times (\mb{p}_{i}-\mb{p}_{i-1})}
{(\mb{p}_{i + 1}-\mb{p}_i) \cdot (\mb{p}_{i}-\mb{p}_{i-1}) }
\label{eq:theta}
\end{equation}
and the arc length is given by
\begin{equation}
\rho_i = (d/2)\cot(\theta_i/2)
\label{eq:rho}
\end{equation}

The elastic strain energy in a beam is given by
\begin{equation}
U = \int_0^L \frac{M^2 ds}{2EI} 
\end{equation}
in which $L$ is the length of the beam, $M$ is the bending moment, $E$ is the elastic modulus and $I$ is the area moment of inertia for the cross section. For a beam experiencing pure bending with radius of curvature, $\rho$, the moment $M$ is constant and given by
\begin{equation}
M=EI/\rho
\end{equation}
We combine the equations for the sleeve covering ball $i$ and obtain an expression for its strain energy, $U_{s_i}$,
\begin{equation}
U_{s_i} = \frac{E_i I_i \theta_i}{2 \rho_i}
\end{equation}
Note that the length of the segment is $\rho_i \theta_i$ and that we assumed that the elastic modulus, $E_i$, and second moment of area, $I_i$, are piecewise constant.

The total strain energy in the sleeve due to bending is obtained by summing over the ball segments 
\begin{equation}
\label{eq:elastic}
U_s = \frac{1}{2}\sum_{i = 1}^n \frac{E_i I_i \theta_i}{\rho_i}
\end{equation}
which is a function of the shape parameters through (\ref{eq:theta}) and (\ref{eq:rho}).

Finally, we write gravitational potential energy as 
\begin{equation}
U_g = \sum_{i = 1}^n \mu_i \mb{g}^T \mb{p}_i,
\end{equation}
where $\mu_i$ is the mass of the $i$th ball and $\mb{g}$ as the gravitational acceleration vector.

By summing all the contributions to the energy, we obtain the expression for total potential energy
\begin{equation}
    \label{eq:equilibrium}
    U = U_b + U_e + U_s +U_g.
\end{equation}

We find the robot's shape as set of shape parameters, $(\mb{p}_i^*, \mb R_i^*), \ i \in 2, \ldots, n$, that minimize potential energy, $U$, i.e.
\begin{equation}
\label{eq:statics}
(\mb{p}_i^*, \mb R_i^*), \ i \in 2, \ldots, n = \argmin_{(\mb{p}_i, \mb R_i), \ i \in 2, \ldots, n} U
\end{equation}
under the constrain that the balls are rigid
\begin{equation}
    |\mb{p}_i| > d \ \forall \ i=1,2,\dots,n.
\end{equation}
The problem can be solved as a constrained function minimization using routines such as {\it fmincon} from Matlab (Mathworks, Natick, MA).

\section{Magnetic Actuation System}
\label{sec:actuation}
\begin{figure}
    \centering
    \begin{subfigure}[b]{\columnwidth}
         \centering
         \includegraphics[width=0.8\columnwidth, bb=0 0 250 250]{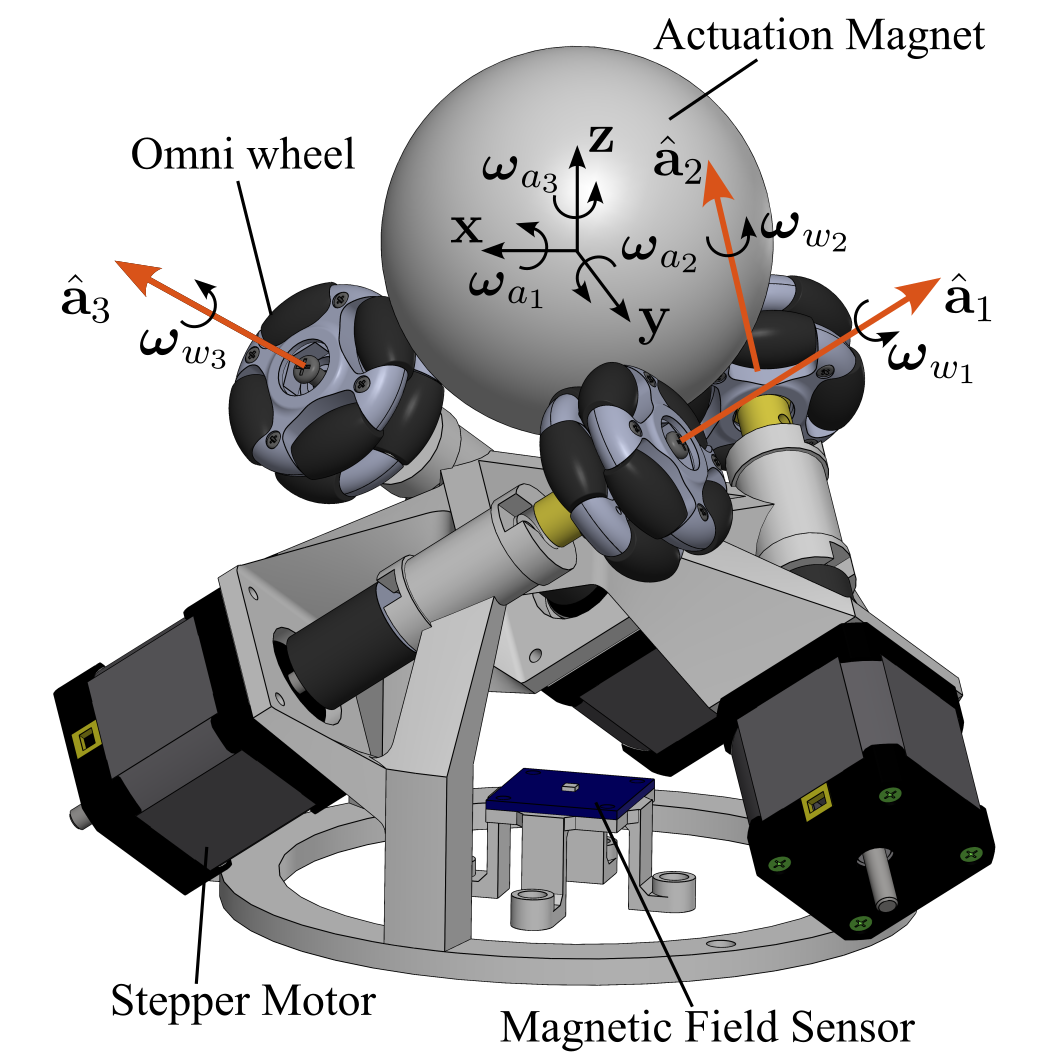}
         \caption{Schematic description of the actuation unit for the rotation of the actuation magnet.}
         \label{subfig:actuation_unit}
    \end{subfigure}
    \hfill
    \begin{subfigure}[b]{\columnwidth}
         \centering
         \includegraphics[width=0.8\columnwidth, bb=0 0 250 170]{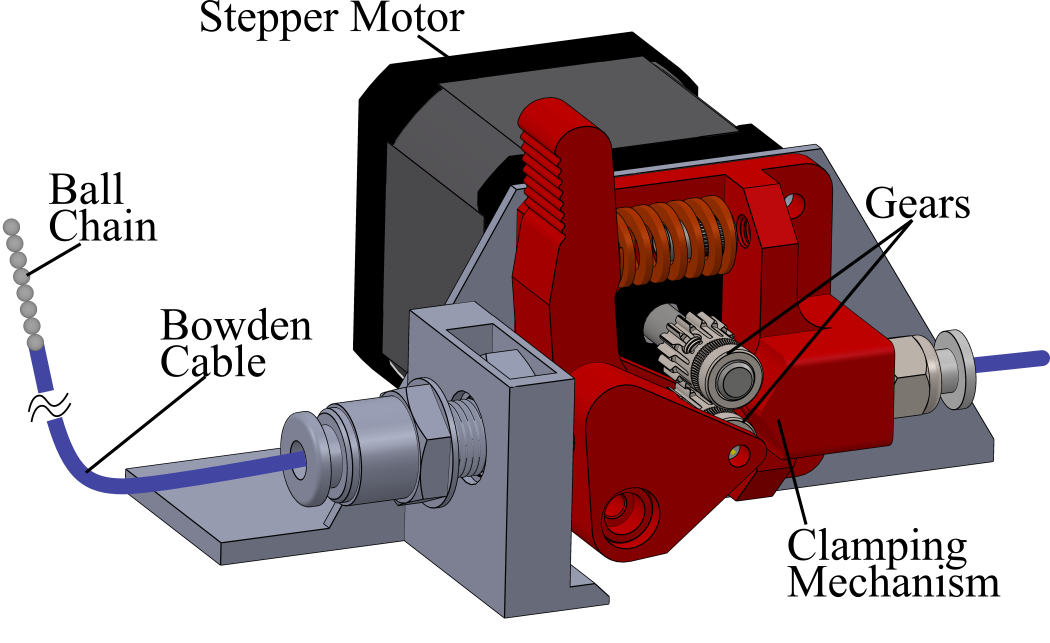}
         \caption{Schematic representation of the feeding mechanism for insertion/retraction of the ball chain.}
         \label{subfig:feeder}
    \end{subfigure}
    \caption{Representation of the actuation method.}
    \label{fig:actuation}
\end{figure}

To assess the concept of Fig. \ref{fig:platform} in which a pair of cart-mounted external magnets control catheter tip motion via pure rotation, the rotation system of Fig. \ref{subfig:actuation_unit} was designed. It uses a configuration of three omni wheels inspired by Wright \emph{et al.} \cite{Wright2015} which are driven by stepper motors (4209L-01S stepper motors (Lin Engineering, USA)) to rotate a spherical magnet (or a spherical shell enclosing a magnet of different shape) which is pressed against the wheels under its gravitational load. The direction of the magnetic field is measured using a three-axis Hall effect sensor (EVB90393; Melexis, Belgium) mounted directly under the spherical magnet.

The wheels are selected to be of the same radius, $\rho_w$, and the radius of the magnet is $\rho_a$. Given the axis of each wheel $\hat{\mb{a}}_i, \ i = 1, 2, 3$, the angular velocity of the wheels $\pmb{\omega}_w \in \mathbb{R}^3$ (each component angular velocity of each wheel) maps to the angular velocity of magnet $\pmb{\omega}_a \in \mathbb{R}^3$ in the global reference frame $\{ \mb{x},\mb{y}, \mb{z}\}$ as \cite{Wright2015}

\begin{equation}
\label{eq:wheel_magn}
    \pmb{\omega}_a = \frac{\rho_a}{\rho_w}
    \left(
    \begin{array}{c}
        -\hat{\mb{a}}^T_1 \\
        -\hat{\mb{a}}^T_2 \\
        -\hat{\mb{a}}^T_3 
    \end{array}
    \right) \pmb{\omega}_w = \eta \mb A \pmb{\omega}_w.
\end{equation}
For the design implemented here, $\mb A$ is given by 
\begin{equation}
    \mb A = \left(
    \begin{array}{ccc}
        0.82 & 0 & -0.58\\
        -0.41 & 0.71 & -0.58 \\
        -0.41 & -0.71 & -0.58
    \end{array}
    \right).
\end{equation}

The ratio $\eta$ can be used to fine tune the rotation accuracy. In the system used for our experiments, in Section \ref{sec:experiments}, we use wheels of radius $\rho_w = 48$mm and a magnet with effective radius, $\rho_a = 43.5$mm. 

To control the inserted catheter length, a feeding system comprised of a Bowden cable extrusion mechanism (NEMA17 stepper motor; STEPPERONLINE Inc., USA) was used as also shown in Fig. \ref{subfig:feeder}. As a cable we used a $\sim 2$m length of PLA filament (HATXHBOX - White PLA, diameter 1.75mm) with the ball chain attached to its tip. A spring-loaded clamping mechanism allows a set of two gears to clamp the cable and transfer the rotation of the stepper into feeding motion.

\section{Magnetic Field Control}
\label{sec:control}
The control of the actuation system consists of two modalities: (i) teleoperation, i.e. the clinician inputs a motion of the chain's tip using a joystick which is mapped to the actuation unit; (ii) axis reconfiguration which, at the beginning of the operation or when called by the clinician, aligns the the magnet with its neutral axis. We describe these modalities in the following.

\subsection{Teleoperation}
\label{sub:teleoperation}
The operator was provided with a joystick (see Fig. \ref{fig:robot_manual_setup} in Section \ref{sec:experiments} for details) which maps onto the rotation of the actuating magnet $\pmb \omega_a$ in global reference frame, $\{\mb x, \mb y, \mb z\}$ in Fig. \ref{subfig:actuation_unit}. A complete control scheme which maps the operator's input into chain tip motion would require solving the model in Section \ref{sec:ball_chain} iteratively. However, here we use a simplified approach, based on the observation that the tip of chain always aligns with the applied magnetic field. 

\begin{figure}
    \centering
    \begin{subfigure}[b]{\columnwidth}
         \centering
         \includegraphics[width=\columnwidth]{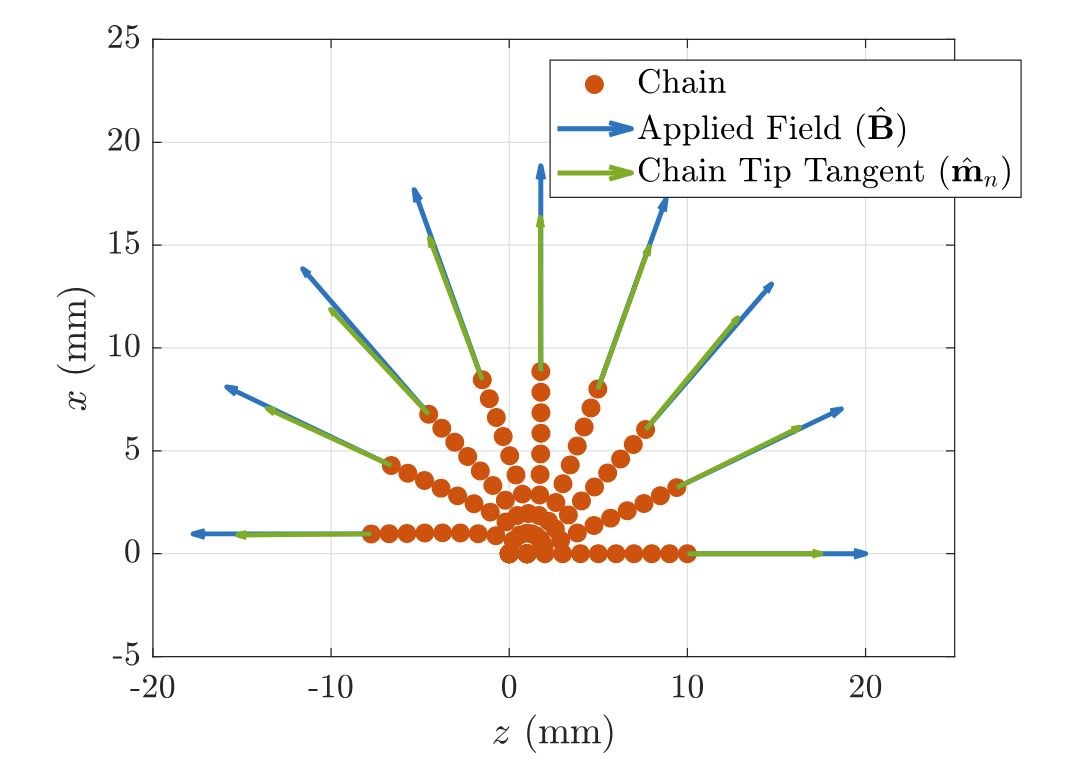}
         \caption{Shape results of the model in Section \ref{sec:ball_chain} for a chain of 10 balls while applying magnetic field of intensity 23mT rotating from 0$^\circ$ to 180$^\circ$, with increments of 22.5$^\circ$.}
         \label{subfig:model_results}
    \end{subfigure}
    \hfill
    \begin{subfigure}[b]{\columnwidth}
         \centering
         \includegraphics[width=\columnwidth]{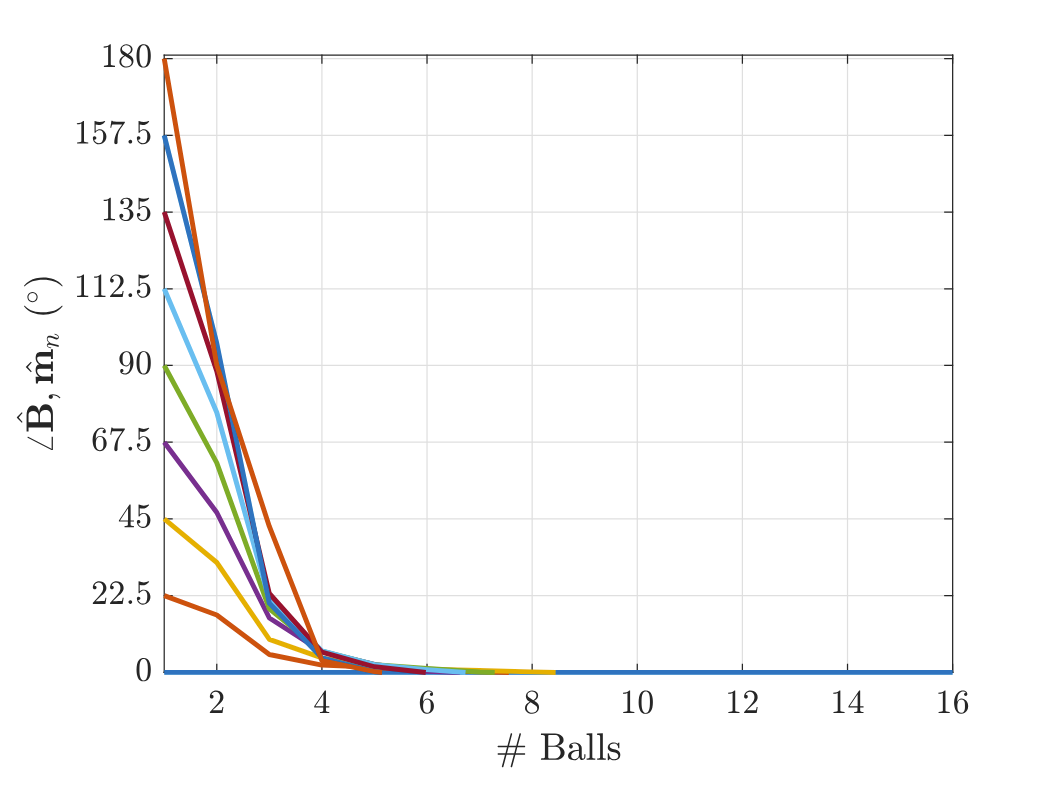}
         \caption{Angle between applied field and chain tip tangent for rotation field from 0$^\circ$ to 180$^\circ$, with increments of 22.5$^\circ$ and varying chain length.}
         \label{subfig:model_error}
    \end{subfigure}
    \caption{Simulation results showing the alignment of the chain's tip with the applied magnetic field.}
    \label{fig:model_simulation}
\end{figure}

To show this, we report the results of the simulated model from Section \ref{sec:ball_chain} in Fig. \ref{fig:model_simulation}. We modelled a chain extending from 1 to 16 balls - same length used in the experiments in Section \ref{sec:experiments} - and applied a rotating magnetic field of intensity 23mT, i.e. the one computed for a magnet of the size. The field direction $\hat{\mb B}$ was rotated from angle $0^\circ$, aligned with the direction of insertion of the chain, to $180^\circ$, anti-parallel to the ball chain. The magnetic dipole direction of the ball at the tip of the chain - i.e. $n$th ball - $\hat{\mb m}_n$ indicates the chain's approach direction. 

In Fig. \ref{subfig:model_error}, we report the angle between $\hat{\mb{B}}$ and $\hat{\mb m}_n$ ($\angle \hat{\mb{B}}, \hat{\mb m}_n$) for different orientations of $\hat{\mb{B}}$ and varying the extension length of the ball chain ($\mb{x}$ axis). We see that, for a ball chain of length greater than 10 balls (represented in Fig. \ref{subfig:model_results}), the angle between its tangent ($\hat{\mb m}_n$) and the applied field is less than 1.4$^\circ$ (0.8\%). 

In the experiments in Section \ref{sub:robotic_vs_manual}, where we show navigation inside the heart, 10 balls is the minimum chain length to reach every target. Therefore, we can confidently infer that in the operative conditions the chain tip can be assumed to align with the magnetic field. 

Using this simplification, a differential rotation of the actuating magnet in Fig. \ref{subfig:actuation_unit} maps into a differential motion of the chain tip's direction. Therefore, the joystick input directly maps in the desired motion of the chain's tip. Using visual feedback, the operator can easily adjust its tip position moving the magnet in its global frame. This is shown in the experiments in Section \ref{sub:robotic_vs_manual}.  

\subsection{Axis Reconfiguration}
The magnet's actuation method proposed can only control the magnet's angular velocity and not its absolute angular position, since there might be possible sliding between magnet and wheels. To be able to reposition the magnet to a neutral position, we use a closed loop control method based on the feedback from a three-axes magnetic field sensor (see Fig. \ref{subfig:actuation_unit}) \cite{Wright2015}.

The actuation magnet's dipole direction ($\hat{\mb{m}}_{a_m}$) can be found from the field recorded by the sensor ($\mb{B}_m$), by inverting the equations of the magnetic dipole (see (\ref{eq:dipole}) in Section \ref{sec:ball_chain}) as 
\begin{equation}
    \hat{\mb{m}}_{a_m} = \frac{4 \pi D^3}{\mu_0 |\mb{m}_{a_m}|} \left(3 \mb{e}_3 \mb{e}^T_3 - I \right) \mb{B}_{a_m}.
\end{equation}
Here $D$ is the distance between the magnet's center and the field sensor; notice that the actuation magnet's dipole intensity ($|\mb{m}_{a_m}|$) is known from its datasheet. In our platform, the field sensor is positioned along the global $\mb z$ axis in negative direction (see Fig. \ref{subfig:actuation_unit}). 

To aid the operator, it may be necessary that the magnet is re-positioned so that the field aligns with the chain's feeding direction (neutral configuration) at the beginning and, in some cases, during the operation. This way, as we introduce it, the chain would follow the direction of feeding and be a compass for the operator's actions. We call the neutral direction for the magnet $\hat{\mb{m}}_{a_c}$ and implemented a initialization/reconfiguration routine which implements the alignment of $\hat{\mb{m}}_{a_m}$ with $\hat{\mb{m}}_{a_c}$. When the reconfiguration callback is called, the wheels' angular velocity is actuated to follow the control law

\begin{equation}
    \pmb{\omega}_w = K \eta^{-1} A^{-1}(\hat{\mb{m}}_{a_m} \times \hat{\mb{m}}_{a_c}),
\end{equation}
until convergence to $\hat{\mb{m}}_{a_m} \times \hat{\mb{m}}_{a_c} \approx 0 \Rightarrow \hat{\mb{m}}_{a_m} \approx \hat{\mb{m}}_{a_c}$, for $K\in \mathbb{R}^+$.

\section{Experiments}
\label{sec:experiments}
In the following, we report experimental validation of the proposed platform for magnetically-enabled robotic navigation for cardiac ablation. 

In all scenarios, the ball chain was composed of 16 Nickel-Plated magnetic balls (N52, diameter: 3.175mm) and the magnet used for actuation a N52 cylindrical magnet of diameter 76.2mm and thickness 38.1mm. When used, the ball chain was embedded in a soft silicone sleeve made of Dragonskin 10NV (Smooth-On, Inc., USA) of ID 3mm and OD 3.5mm. In Sections \ref{sub:workspace}, we show how this sleeve does not significantly affect the workspace of the ball chain alone. Also, in Section \ref{sub:ablation} we demonstrate that it can stand heating related to performing ablation. The chain's length was selected so that it could reach any target location in the heart, in the experiments reported in Section \ref{sub:robotic_vs_manual}.

\subsection{Force Evaluation}
\label{sub:force}
\begin{figure}
    \centering
    \begin{subfigure}[b]{\columnwidth}
         \centering
         \includegraphics[width=0.75\columnwidth, bb=0 0 160 170]{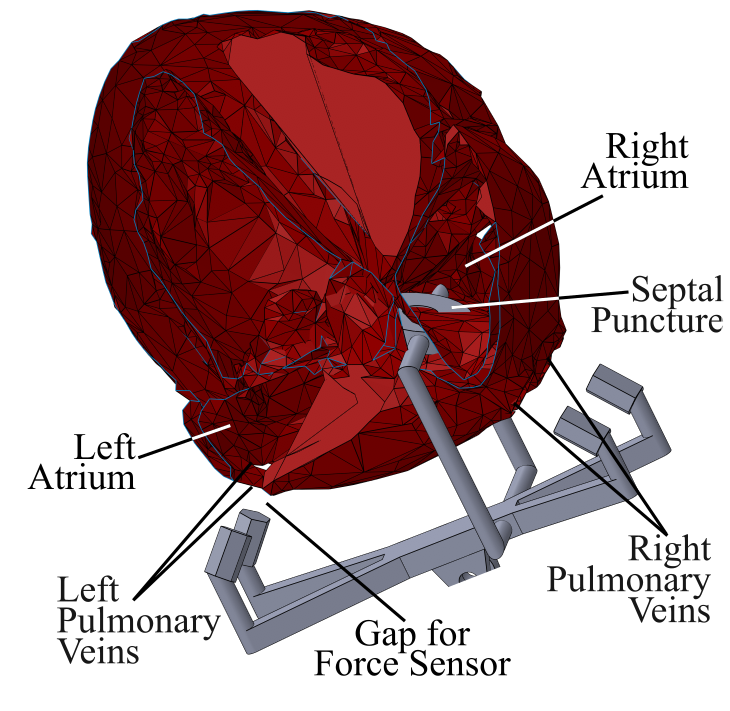}
         \caption{Representation of 3D printed stand abstracted from heart anatomy.}
         \label{subfig:force_setup_abstract}
    \end{subfigure}
    \hfill
    \begin{subfigure}[b]{\columnwidth}
         \centering
         \includegraphics[width=\columnwidth, bb=0 0 250 170]{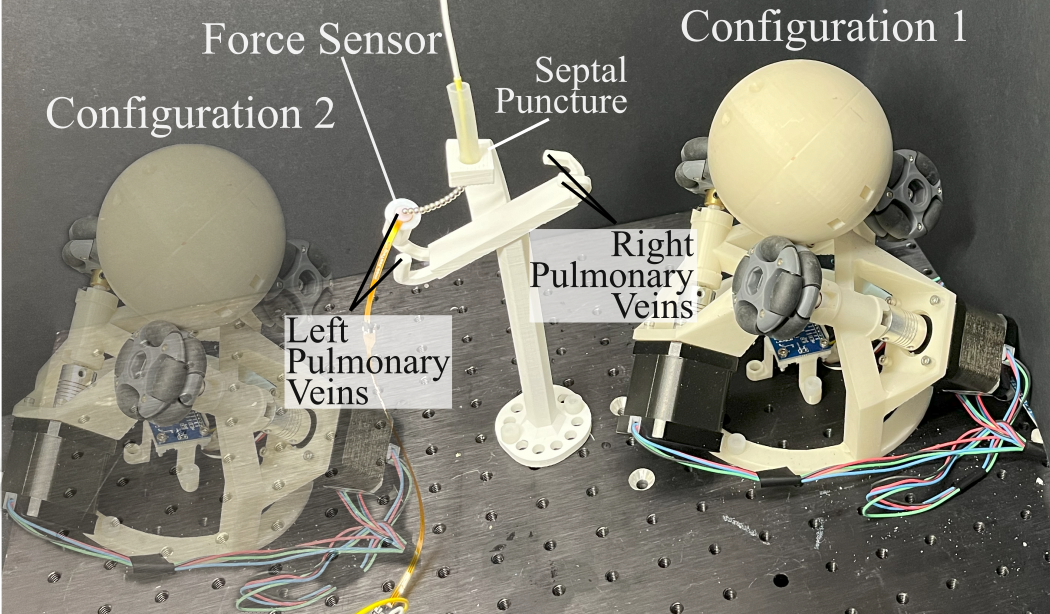}
         \caption{Setup for analysis of the force applied by a ball chain, using the proposed actuation unit. Configuration 1 and 2 indicate two independent positions of the actuation unit used to gather force data.}
         \label{subfig:force_setup_pic}
    \end{subfigure}
    \caption{Setup for analysis of force exerted by the ball chain on targets on pulmonary veins.}
    \label{fig:force_setup}
\end{figure}

We 3D printed a stand (see Fig. \ref{fig:force_setup}) with an insertion point (septal puncture) and the location of the four pulmonary veins (left and right). The locations were extracted from the same anatomical phantom which we used in the experiments in Section \ref{sub:robotic_vs_manual}, as depicted in Fig. \ref{subfig:force_setup_abstract}.

For the purpose of understanding the forces involved, we positioned a force sensor on each pin representing a pulmonary vein. This is used as an approximation of the forces applied in the area, which would normally be around the veins. We positioned our actuation unit at a distance of 16.51cm from the septal puncture which is aligned with the center of the actuating permanent magnet. 

We first collected the force data on all four pulmonary veins with the unit on the right side of the heart phantom (Configuration 1 in Fig. \ref{fig:force_setup}) and then recollected the data with the unit to the opposite side (Configuration 2 in Fig. \ref{fig:force_setup}).

Under the assumption that the ball chain has minimal influence in deflecting the applied magnetic field, we use the superimposition principle to understand the total force that a combination of two actuation units would generate. We report the value of the total force, along with the measured values found in Configuration 1 and 2, in Table \ref{tab:force_meas}. Since chain bending contributes to the applied force, we generally measure higher forces with the sleeve. The stiffening effect, which does not negatively impact the workspace significantly, has a positive effect on the applied forces.

\begin{table}[]
\centering
\caption{Force measurements by the ball chain with and without sleeve in grams.}
\label{tab:force_meas}
\begin{tabular}{l|lll||lll|}
\cline{2-7}
                                    & \multicolumn{3}{l||}{\textbf{Without Sleeve}}                                                       & \multicolumn{3}{l|}{\textbf{With Sleeve}}                                                          \\ \cline{2-7} 
                                    & \multicolumn{1}{l|}{\textbf{Conf. 1}} & \multicolumn{1}{l|}{\textbf{Conf. 2}} & \textbf{Total} & \multicolumn{1}{l|}{\textbf{Conf. 1}} & \multicolumn{1}{l|}{\textbf{Conf. 2}} & \textbf{Total} \\ \hline
\multicolumn{1}{|l|}{\textbf{LSPV}} & \multicolumn{1}{l|}{20.4}               & \multicolumn{1}{l|}{91.8}               & \textit{112.2} & \multicolumn{1}{l|}{30.6}               & \multicolumn{1}{l|}{107.1}               & \textit{137.7} \\ \hline
\multicolumn{1}{|l|}{\textbf{LIPV}} & \multicolumn{1}{l|}{20.4}               & \multicolumn{1}{l|}{86.7}               & \textit{107.1} & \multicolumn{1}{l|}{30.6}               & \multicolumn{1}{l|}{102.0}              & \textit{132.6} \\ \hline
\multicolumn{1}{|l|}{\textbf{RSPV}} & \multicolumn{1}{l|}{81.6}               & \multicolumn{1}{l|}{40.8}               & \textit{122.4} & \multicolumn{1}{l|}{81.6}               & \multicolumn{1}{l|}{51.0}               & \textit{132.6} \\ \hline
\multicolumn{1}{|l|}{\textbf{RIPV}} & \multicolumn{1}{l|}{81.6}               & \multicolumn{1}{l|}{40.8}               & \textit{122.4} & \multicolumn{1}{l|}{81.6}               & \multicolumn{1}{l|}{51.0}               & \textit{132.6} \\ \hline
\end{tabular}
\end{table}

To understand the dimensions of our actuation units for clinical applications, we need to consider how force scales as we position our actuation units outside of the body of the patient.

\begin{figure}
    \centering
    \includegraphics[width=\columnwidth, bb=0 0 250 120]{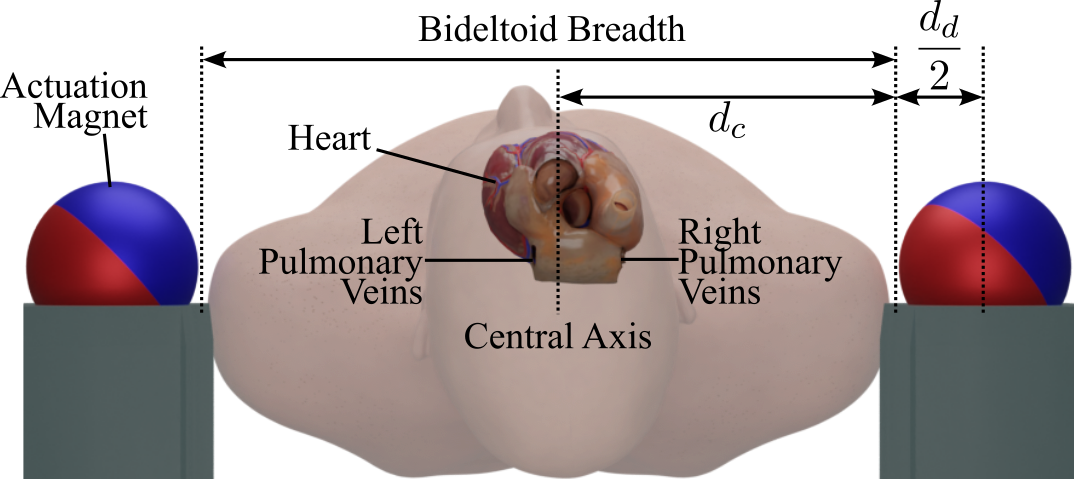}
    \caption{Schematic representation of the platform's and patient's dimensions.}
    \label{fig:platform_project}
\end{figure}

We considered that the units where to be placed on each side of the chest of the patient (as indicated in Fig. \ref{fig:platform}). Based on \cite{Gordon2014} we found a maximum bideltoid breadth (total width considering arms) of 59.30cm as the 99th percentile of men. We define $d_c = 59.30/2 = 29.65$cm as the distance between the heart and the external side of the patient's arm (see Fig. \ref{fig:platform_project}) and $d_d$ as the desired diameter of the magnet. Thus, the heart-magnet distance must be of at least $d = d_c + d_d/2$ so that the magnet is outside the patient's body. 

Another parameter to consider is how force scales with the size of the magnetic balls in the chain. The magnets we currently use, since readily found on the market, are balls of diameter 3.175mm. However, the end goal is to fabricate magnetic catheters of 8Fr (2.67mm diameter) - the same diameter as existing ablation catheters. If we consider a sleeve of approximately 0.5mm thickness, as the one proposed in the present paper, we should select balls of diameter approximately  2.17mm. This corresponds to scaling the volume, hence the dipole intensity of the chain, by a factor $\alpha = 0.32$.

From the equation of the dipole (\ref{eq:dipole}), we can write the force applied by an external permanent magnets of magnetic dipole $\mb m_a$, on an element of magnetization $\mb m_i$ in position $\mb p_i$ with respect to the center of the magnet as $\nabla (\mb m_i \cdot \mb B(\mb p_i, \mb m_a))$. By manipulating the expression, we can find the force intensity as
\begin{equation}
\label{eq:force_intens}
    |\mb f | = \frac{3 \mu_0 \ |\mb m_i| \ |\mb m_a|}{4 \pi d^4}.
\end{equation}
where $d = |\mb p_i|$.
As in Section \ref{sub:teleoperation}, here we assume that the chain aligns with the applied magnetic field.

We assume that the superimposition principle applies locally, on any point far from each actuating magnet of magnetic dipole intensity $|\mb m_a|$; hence, we consider double the force in (\ref{eq:force_intens}) as applied by two actuating magnets. In this analysis, $\mb m_i$ only approximates the dipole of the overall chain, and is constant and independent from moving the external magnets away (changing $d$) or changing their dipole intensity $|\mb m_a|$. 

We write the respective measured and the desired force as
\begin{equation}
\label{eq:meas_force}
    |\mb f_m | = \frac{3 \mu_0 \ |\mb m_m| \ |\mb m_a|}{4 \pi d^4}
\end{equation}
and
\begin{equation}
\label{eq:des_force}
    |\mb f_d | = \frac{3 \mu_0 \ |\mb m_d| \ |\mb m_a|}{4 \pi d^4}.
\end{equation}
where the subscript $m$ refers to the values measured in our experiments and $d$ for the desired values - i.e. the ones that would allow appropriate clinical application of the proposed platform.

Given the reduction the volume of each ball in the chain by $\alpha$ we can write 
\begin{equation}
\label{eq:dipole_intens}
     \alpha |\mb m_m| = |\mb m_d|
\end{equation}
and, by combining (\ref{eq:dipole_intens}), (\ref{eq:des_force}) and (\ref{eq:meas_force}), we obtain
\begin{equation}
\label{eq:total_force}
     \frac{4 \pi d_m^4 \alpha |\mb f_m |}{3 \mu_0 \ \ |\mb m_{a_m}|} = \frac{4 \pi d^4 |\mb f_d |}{3 \mu_0 \ \ |\mb m_{a_d}|},
\end{equation} 
For the measured force, we use the worst case force from the chain with the sleeve $|\mb f_m| = 132.6$g, from Table \ref{tab:force_meas}, while the desired one is $|\mb f_d| = 10.0$g. The dipole intensity can be computed, for both measurement and desired, as $|\mb m_{a_m}| = B_r V_m/\mu_0$ and $|\mb m_{a_d}| = B_r V_d/\mu_0$, respectively; here $B_r$ is the residual flux density and $V_m$, $V_d$ the volume of the magnet used for the measurements and the desired magnet, respectively.

If we substitute these values into (\ref{eq:total_force}) and manipulate the equation, we obtain that
\begin{equation}
    \alpha |\mb f_m| d_m^4 V_d = |\mb f_d| d^4 V_m 
\end{equation}
If, for simplicity's sake, we consider using a spherical a spherical magnet for actuation, we obtain the following expression
\begin{equation}
    \alpha \frac{4}{3}\pi|\mb f_m| d_m^4 \left(\frac{d_d}{2}\right)^4 - |\mb f_d| \left(d_c + \frac{d_d}{2}\right)^4V_m = 0 
\end{equation}
which we solve with respect to $d_d$ to find the diameter of the magnet for clinical application. We obtain a magnet of diameter $d_d = 119$mm and mass of approximately $6.6$kg. Two of these magnets, one on each side of the patient, provide a tip force of the ball chain of 10g.

For magnets of this scale, a simple mounting system would be required. Given their small diameter, they would have a limited footprint when positioned on each side of the patient. Also, their light weight allows easy mounting on a movable cart - which can be brought in when required, and does not involve changes in the current clinical workflow.  With respect to safety, the force between the two magnets when positioned on each side of the patient is about 500 gram-force, which is completely safe.

\subsection{Workspace Evaluation}
\label{sub:workspace}
\begin{figure}[t]
\includegraphics[width=\columnwidth, bb=0 0 250 200]{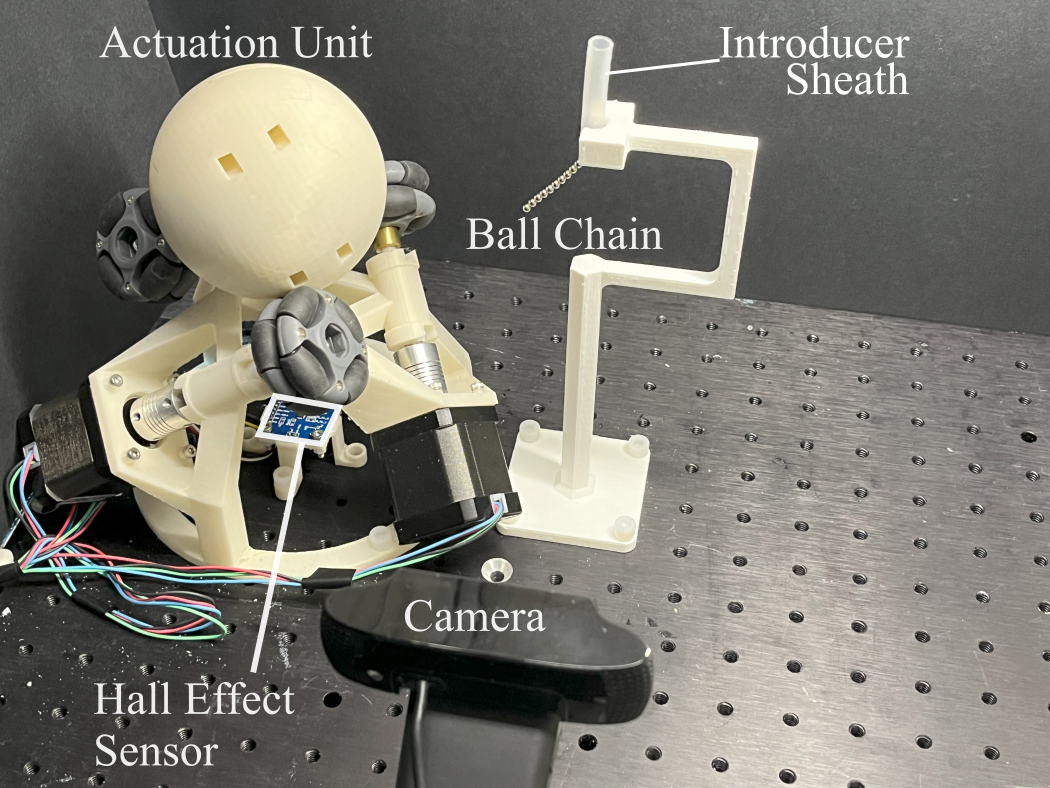}
\caption{Setup for analysis of the workspace covered by a ball chain with and without sleeve.}
\label{fig:workspace_setup}
\end{figure}

The workspace of a ball chain was evaluated with and without a soft sleeve to understand whether or not the sleeve would affect the bending capabilities of the ball chain. The ball chain was clamped via a 3D printed stand at a distance of 16.51cm from the actuation unit (see Fig. \ref{fig:workspace_setup}). While the proposed actuation system is composed of two external magnets placed on each side of the patient, these tests employed a single magnet. Consequently, the results reported here underestimate the workspace that could be reached with two external magnets. The point of insertion of the chain was aligned with the center of the actuating permanent magnet. The magnet was rotated to generate a rotating field which reflects into the full range of motion of the chain.

\begin{figure}[t]
\includegraphics[width=\columnwidth, bb=0 0 250 200]{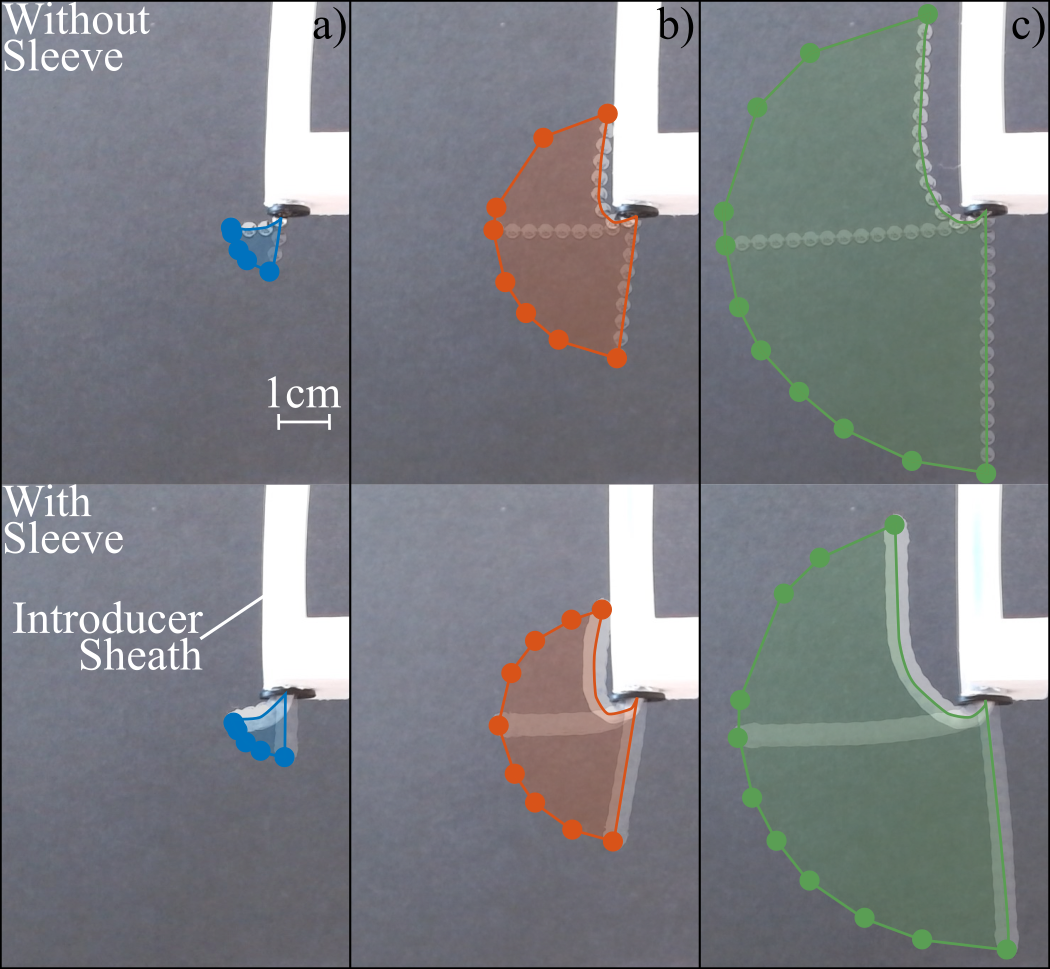}
\caption{Results of workspace analysis with and without sleeve. a) 4 balls inserted; b) 9 balls inserted; c) 16 balls inserted.}
\label{fig:workspace_results}
\end{figure}

A camera was used to capture the configuration of the chain. In Fig. \ref{fig:workspace_results}, we report the results obtained with chain of different lengths: 4, 9 and 16 balls. For practical use in ablations, as reported in Section \ref{sub:robotic_vs_manual}, we require a maximum of 16 balls (47.625mm length). The chain was introduced into the workspace via a tube (introducer sheath in Fig. \ref{fig:workspace_setup} and \ref{fig:workspace_results}) which served to secure the length of the chain which was not actively actuated. This section of the chain, during the experiments was constrained inside the sheath.

We collected a set of positions of the chain's tip, indicated by a solid dot in Fig. \ref{fig:workspace_results}, by rotating the applied magnetic field.

By intersecting the workspace covered at different lengths, highlighted in Fig. \ref{fig:workspace_results}, we computed the area of the analyzed workspace (half the total workspace), resulting in 3345mm$^2$ without sleeve and 2939mm$^2$ with sleeve. Hence, we see only a 12\% reduction of the total workspace by adding a soft sleeve to the ball chain, which does not prevent reaching extreme locations in the anatomy. This can be seen in Fig. \ref{fig:workspace_results}c, where the ball chain with sleeve shows very minimal difference in reach compared to the one without. We measured a tip-to-tip distance of approximately 8mm - only 16\% of total length of the chain. 

\subsection{Model Validation}
\label{sub:model_validation}
\begin{figure}[t]
    \centering
    \includegraphics[width=\columnwidth, bb=0 0 250 250]{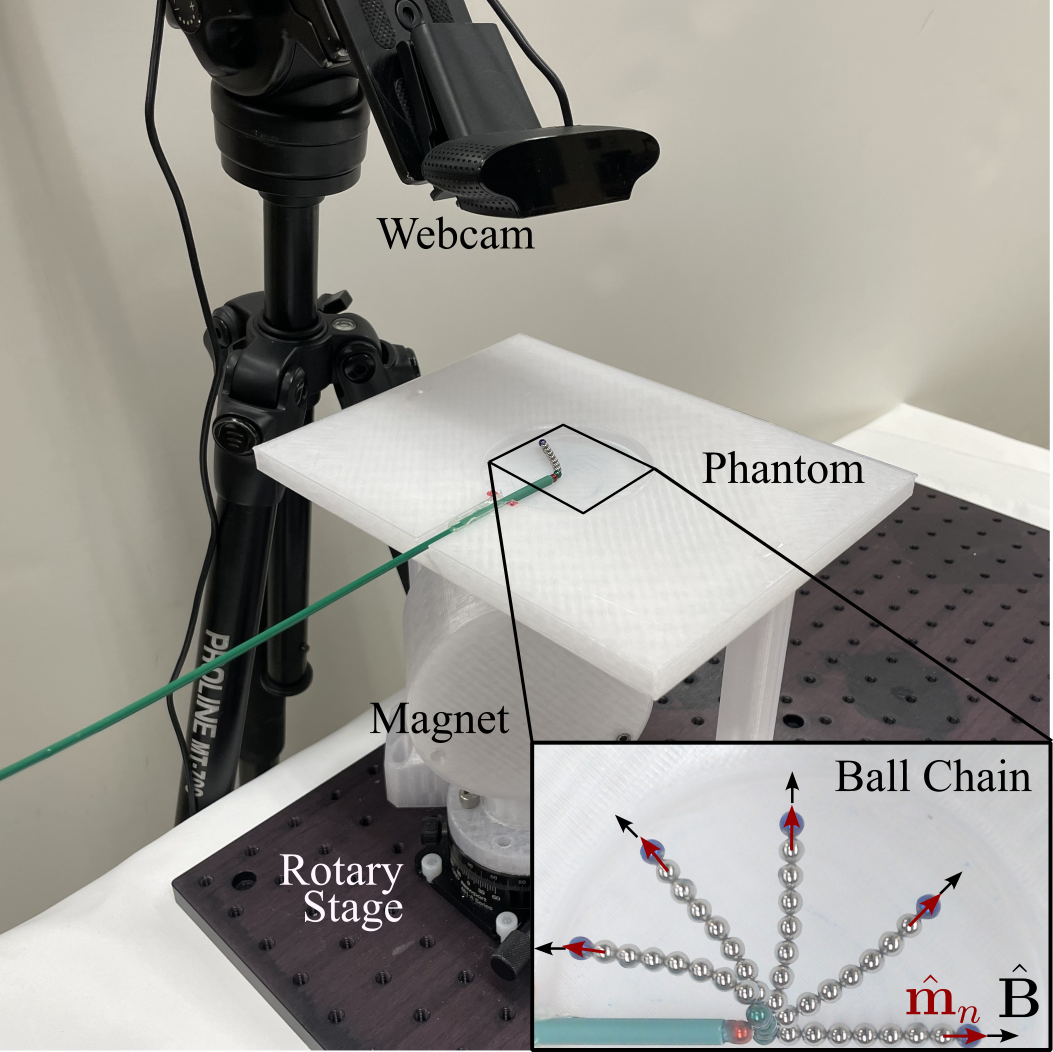}
    \caption{Experimental setup for model validation.}
    \label{fig:model_setup}
\end{figure}

\begin{figure}[t]
    \centering
    \includegraphics[width=\columnwidth, bb=0 0 320 320]{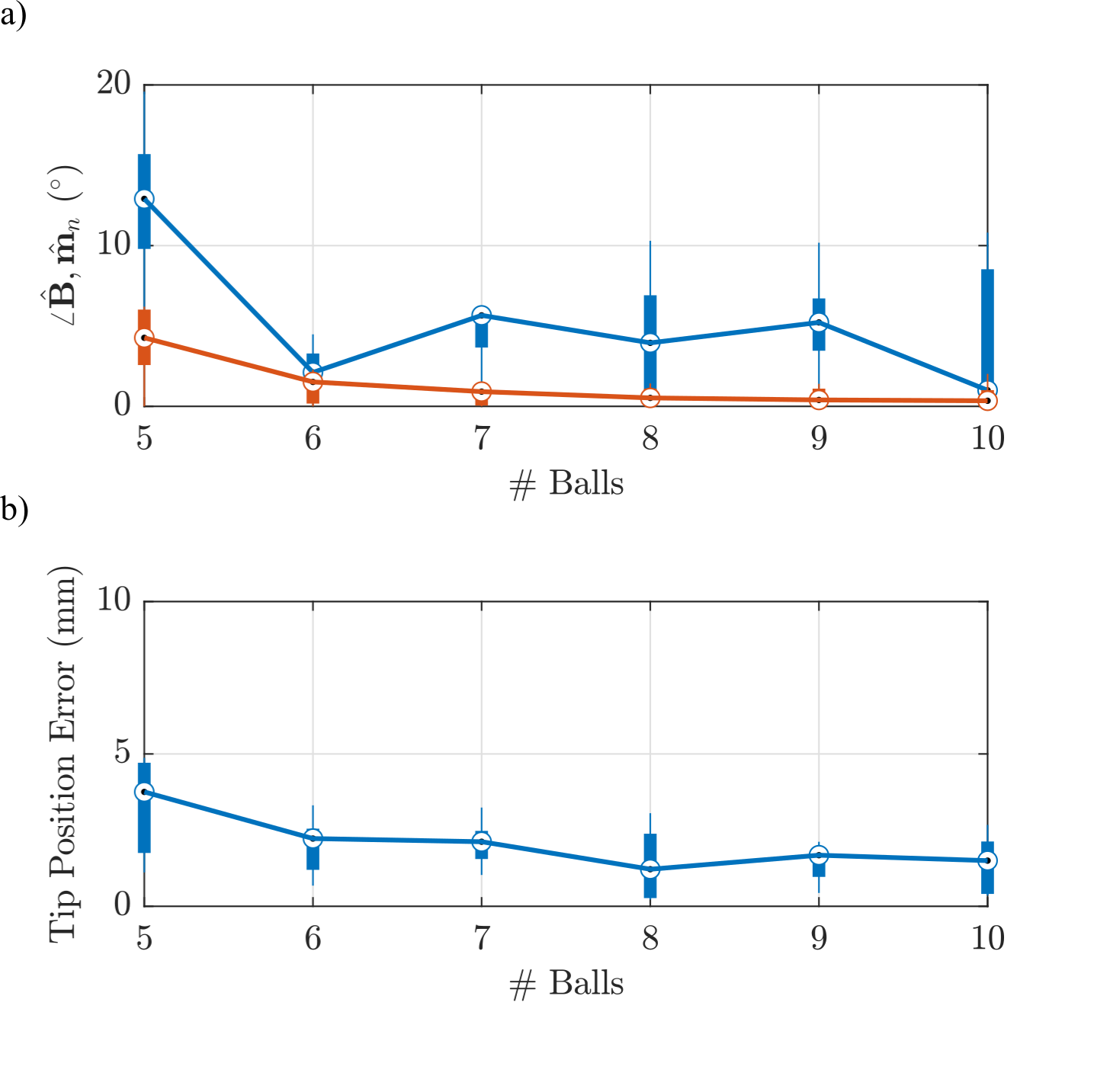}
    \caption{Model validation experiment. a) Angle between external magnetic field and tip tangent direction versus number of balls in chain. b) Tip position error versus number of balls in chain. Circles indicate median, and the bottom and top edges of the box indicate the 25th and 75th percentiles, respectively. The whiskers extend to the most extreme data points.}
    \label{fig:model_results}
\end{figure}

We validated the kinematic model of Section \ref{sec:ball_chain} using the experimental setup in Fig. \ref{fig:model_setup}. The N52 actuation magnet described above was mounted on a single-axis rotary stage so that its field orientation could be precisely measured. Ball chains were positioned on a flat 3D printed platform to eliminate the effect of gravity. For chain lengths varying between 5 and 10 balls, a magnetic field of intensity 23mT was applied in the directions $\{0, 45, 90, 135, 180\}$ degrees. This field intensity is the minimum applied by the permanent magnets required in our application at the appropriate distance from the heart; see Section \ref{sub:force}.

A webcam (C920, Logitech, US) pointing towards the phantom was used to track the ball chain orientation and tip position. For tracking purposes, three balls of the chain were colored in red, green, and blue. Experiments were performed without the sleeve, as it was shown to have negligible effect on chain bending. The Matlab function {\it imfindcircles} was used to find the position of the spheres in the chain; its tip direction, $\mathbf{\hat m_n}$, was calculated as the vector connecting the centers of the two distal balls in the chain.

Figure \ref{fig:model_results} compares experimental and model predictions of tip tangent direction and tip position. Part (a) is a boxplot of the angle between the applied field and the tip tangent direction ($\angle \mathbf{\hat B}, \mathbf{\hat m_n}$) considering the five measured magnetic field angles. This plot confirms the model prediction that, as the length of the chain increases, the tip of the chain aligns with the applied field (orange curve). Experimentally, the chain tip tangent also aligns with the external field. For chains of 6 or more balls, the median and maximum errors between experimental tip tangent and magnetic field direction are 5$^\circ$ and 8$^\circ$, respectively. This indicates that assuming alignment of the chain with the applied field is an appropriate simplification for robotic control. 

From Figure \ref{fig:model_results}b, the maximum error in tip position for chains of 5 or more balls is less than 5mm and, for 6 or more balls, is less than 3mm. The median errors are 3.7mm for 5 balls and 1.5mm for a chain length of 10 balls. These are less than or comparable to sphere diameter and highlight that our model can accurately predict the chain's configuration. Discrepancies between model and experiment may be due to friction between the chain and the 3D printed platform.

\subsection{Robotic vs Manual Control}
\label{sub:robotic_vs_manual}
\label{sub:robotic_vs_manual}
\begin{figure}[t]
\includegraphics[width=\columnwidth, bb=0 0 250 350]{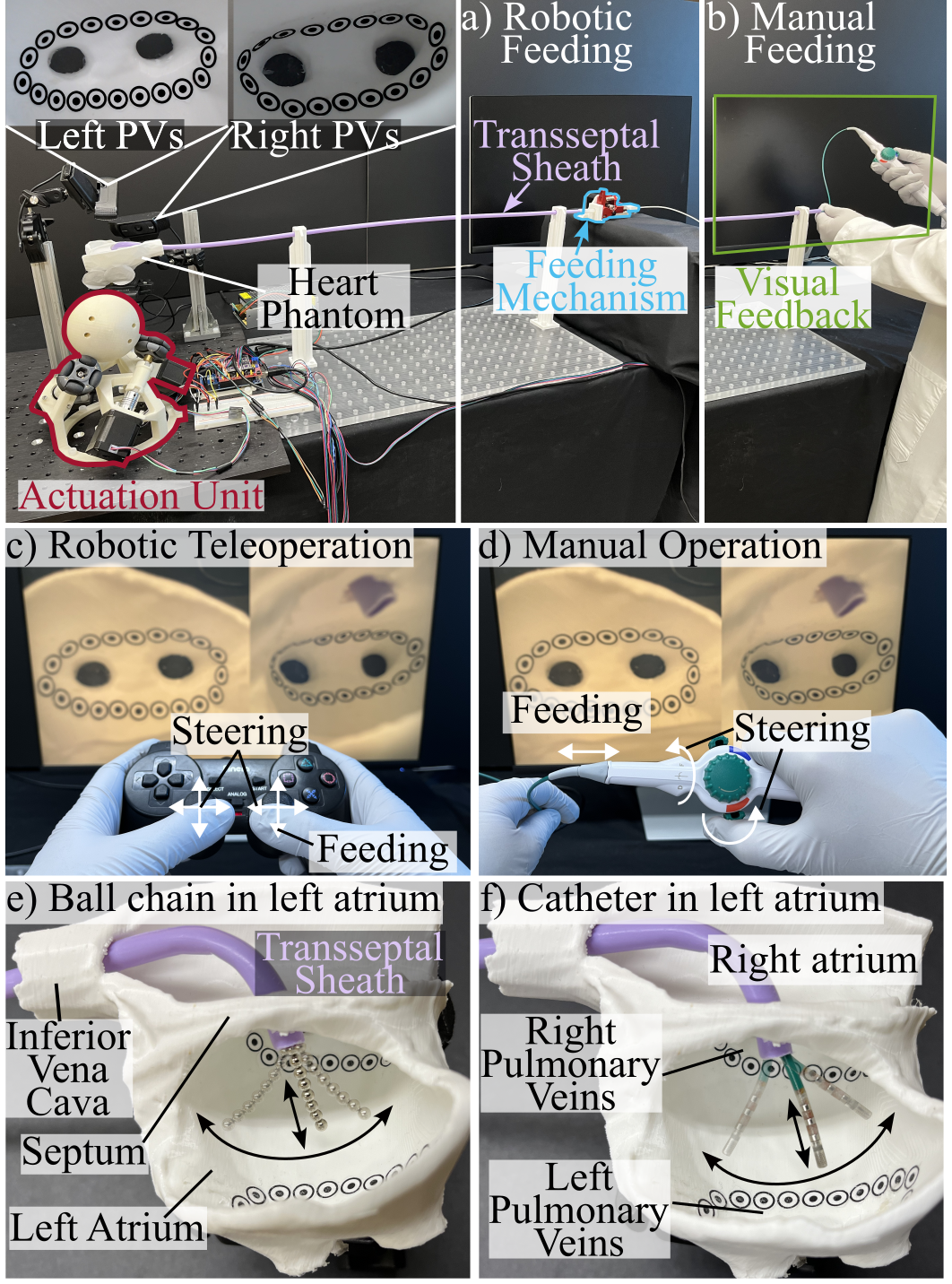}
\caption{Description of the experimental setup for comparative studies between robotic and manual navigation. a) robotic feeding; b) manual feeding; c) robotic teleoperation using joystick; d) manual operation of standard catheter; e) view of the heart with the ball chain; f) view of the heart with the manual catheter. a) \& c) operational mode with robotic platform; b) \& d) operational mode with manual catheters.}
\label{fig:robot_manual_setup}
\end{figure}

To evaluate the performance of the robotic system, we conducted a usability study and compared it with the manual procedure. A novice (first author of the paper) was instructed, by demonstration, in the use of a manual ablation catheter (Tactiflex D-F, Abbott; 8FR diameter) by an expert electrophysiologist (fifth author) and learned using the robotic platform by trials. The novice had no prior experience with any manual or robotic ablation procedures. 

Figure \ref{fig:robot_manual_setup} shows the setup used for both manual operation and robotic teleoperation of the ablation catheters. In both cases a semi-rigid transseptal sheath was introduced in the left atrium (where ablation for atrial fibrillation is mostly performed) via a septal puncture in a 3D printed heart phantom made of PLA material (HATXHBOX - White PLA); the heart anatomy is from the library blenderkit (blenderkit.com) and its anatomical consistency was validated by a expert cardiologists. The heart phantom is depicted in Fig. \ref{fig:robot_manual_setup}e with the ball chain extending from transseptal sheath and Fig. \ref{fig:robot_manual_setup}f with the manual catheter extending in the left atrium. 

The atrial septum was reached by accessing the right atrium through the \gls{ivc}. Extending through the \gls{ivc} towards the femoral vein, the sheath was secured using two 3D printed stands which emulate the effect of vasculature.
The proximal part of the sheath, which would naturally emerge from the femoral vein, was used to introduce both robotic (Fig. \ref{fig:robot_manual_setup}a) and manual (Fig. \ref{fig:robot_manual_setup}b) catheters from this point to the left atrium and reach the pulmonary veins. 

At the point of insertion of the sheath, the manual catheter was introduced via manual pushing and pulling. Twisting the handle and using a tendon pulling mechanism (steering) at the base allowed controlling of the tip to follow the prescribed paths in the heart anatomy, as shown in Fig. \ref{fig:robot_manual_setup}d. 

The actuation system of the robotic platform is described in Section \ref{sec:actuation}. The robotic steering (Fig. \ref{fig:robot_manual_setup}e) and feeding (Fig. \ref{fig:robot_manual_setup}a) was teleoperated using a joystick, as shown in Fig. \ref{fig:robot_manual_setup}c. 

In both cases, the operator used visual feedback provided by two cameras showing the left and right pulmonary veins (PVs), on a computer screen.

The sole magnet used in this paper was placed on the side of the right pulmonary veins at a distance of 10cm from the carina (approximately 17cm from the center of left atrium). We made this choice since this area is the hardest to reach and required more pulling force than in reaching the left pulmonary veins. A complete system would have magnets on both sides to make sure contact is guaranteed everywhere in the crucial areas of the anatomy.

We also avoided to use a sleeve for these experiments, since it was easier to visualize when the naked chain touches the targets from camera views. In fact, the chain has a small deflection which can be noted without the sleeve and used to assess when the target is touched. In Section \ref{sub:workspace}, we demonstrated that the sleeve does not affect the chain's reachable workspace.

We encircled the pulmonary veins with 20 (left veins) and 18 (right veins) printed targets of 5mm diameter to simulate the closed paths required to insulate the veins and re-establish healthy heart beating. Lesions are expected to have an approximate width of 5 to 6mm, hence 5mm lesion-to-lesion distance guarantees we create a closed loop around the veins.
The operator was first trained (10 trials) to touch every one of these targets ordered clockwise starting from the carina's most inferior point as fast as possible, with both platforms. The operator was instructed to touch the center of the target, however the target was considered touched when the catheter/chain touched inside the external ring. In that case, the operator would move to the next target.

\begin{figure}
    \centering
    \includegraphics[width = \columnwidth, bb=0 0 400 320]{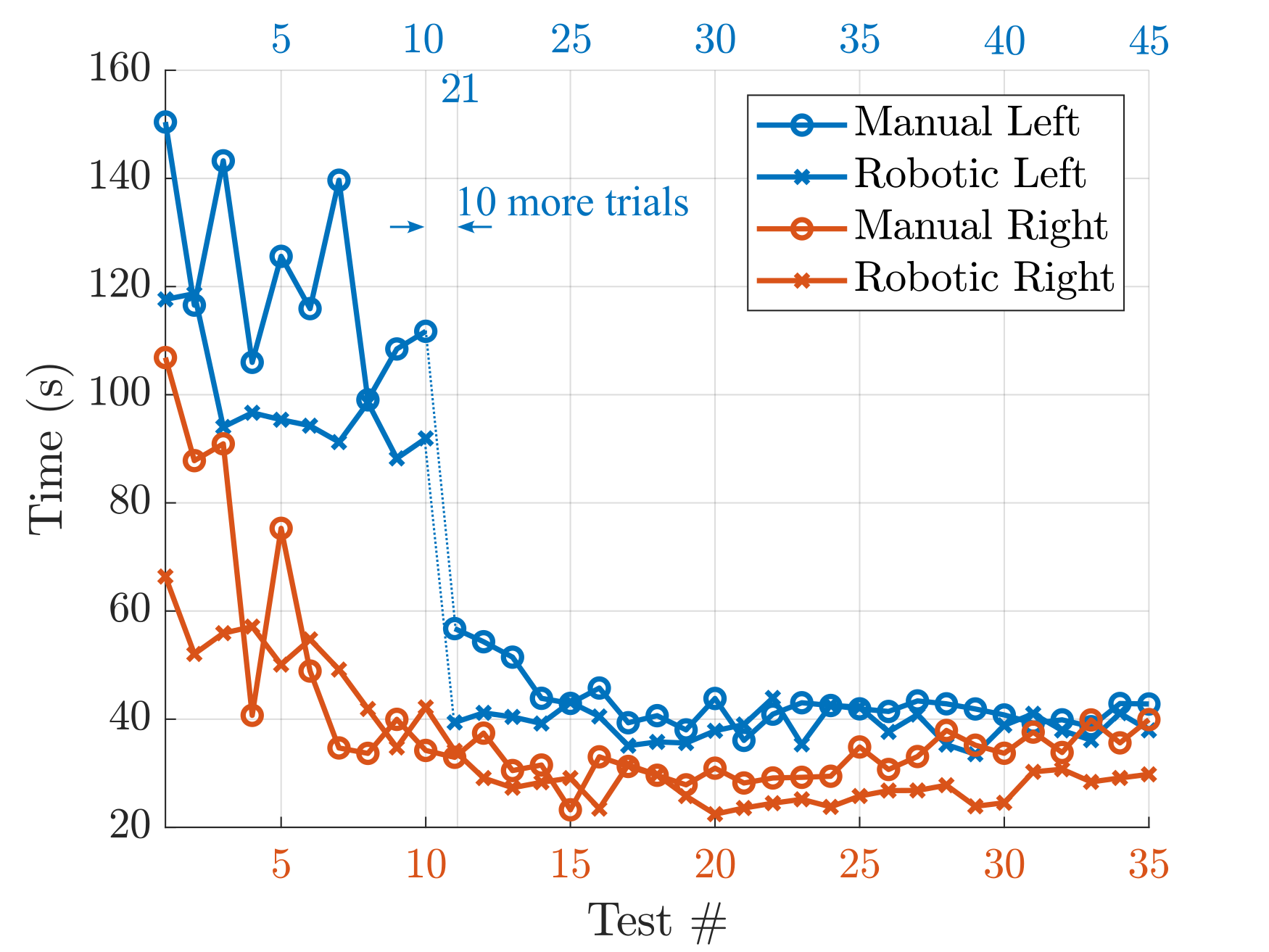}
    \caption{Learning curve for novice operator in navigating manual and robotic ablation catheters around left and right pulmonary veins.}
    \label{fig:learning}
\end{figure}

After learning was ensured (convergence of the learning curve in Fig. \ref{fig:learning}), the operator was tasked to perform 20 trials to make sure task had been learned. Then, we collected 5 trials where we measured both total time and time between consecutive targets. The latter indicates how easily the operator could move to consecutive points with both platforms. The Supplementary Video S1 shows these trials with both robotic and manual tools and underlines the differences in usability and accuracy.

Given that the left pulmonary veins are easier to reach, since directly in front of the septal puncture, the first task (Task I) was to circumnavigate those. As a second task (Task II) the operator was requested to navigate and touch targets around the right pulmonary veins.
In evaluating the operator's learning curve in Fig. \ref{fig:learning}, we notice a jump in the learning of Task I - both manually and robotically. This is because the the operator required more training beyond the expected 10 training trials and 10 more trials where performed before the final tests - without recording - to give the operator more confidence in performing both tasks. This was not required for Task II; even if the task is harder, prior learning from Task I helped familiarizing with the platforms and better learn Task II. 

To make sure that the learning of Task II would not transfer back to performing Task I, we tasked the operator with navigating the left pulmonary veins again - at the end of all the trials. The performance of this further execution of Task I did not outperform the original execution after learning was achieved (see Fig. \ref{fig:learning}). This indicates that the original timings taken while executing Task I were accurate.

In Fig. \ref{fig:learning}, the first trial is about 30s faster with the robotic platform compared to the manual, however, after a total of approximately 30 tests, the performance is comparable and both operations yield a total time of approximately 40s. A similar learning trend can be noticed with Task II, where the first robotic trial is 40s faster than its manual counterpart. In this case, after learning is completed the operator is able to perform the robotic task in 26s, on average, against the 33s in using the manual catheter. 

Counter-intuitively, the harder Task II was completed faster than Task I. This can be understood by noting that the the length of the path around the right pulmonary veins is 95mm while that of the left veins is 115mm. Furthermore, Task II requires less insertion/retraction of the catheter. This enables faster task execution.

\begin{figure}
    \centering
    \begin{subfigure}[b]{\columnwidth}
         \centering
         \includegraphics[width=\columnwidth, bb=0 0 250 70]{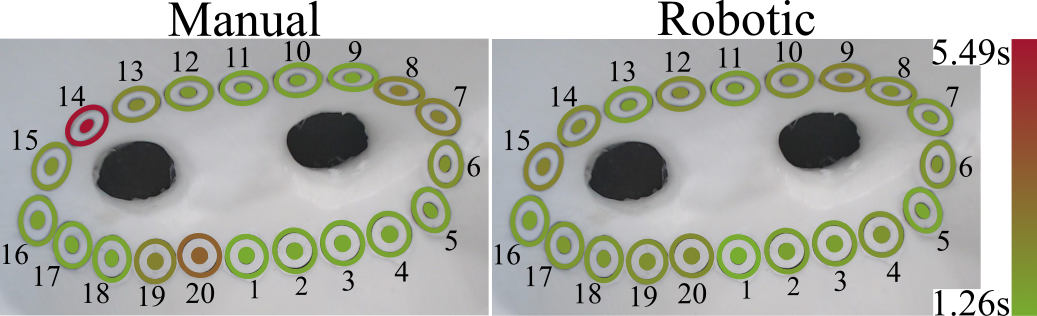}
         \caption{Left pulmonary veins.}
         \label{subfig:target_left}
    \end{subfigure}
    \hfill
    \begin{subfigure}[b]{\columnwidth}
         \centering
         \includegraphics[width=\columnwidth, bb=0 0 250 80]{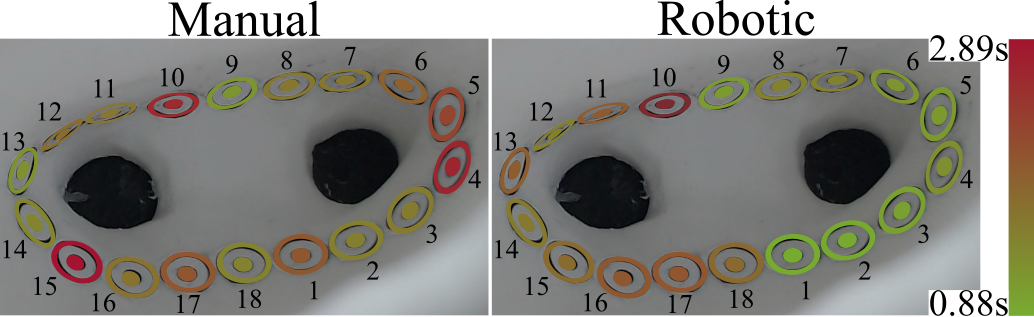}
         \caption{Right pulmonary veins.}
         \label{subfig:target_right}
    \end{subfigure}
    \caption{Evaluation of time between consecutive targets.}
    \label{fig:target}
\end{figure}

Figure \ref{fig:target} shows the average target-to-target timing recorded during the last five repetitions of Fig. \ref{fig:learning}. In this representation, we can appreciate the target locations which are harder to reach. 

For the left pulmonary veins, as shown in Fig. \ref{subfig:target_left}, the transition between target 13 and 14 is the hardest with the manual catheter, with average time of 5.49s. The reason is that after an approximately 180$^\circ$ turn of the handle (from target 1 to 13) reaction force builds between the catheter and the sheath. Therefore, the operator can either keep turning with lower controllability or use the opposite cable by turning the handle back to a 0$^\circ$ twist. Both these operations are time consuming.

In contrast, this issue does not arise with the robotic platform and the the left veins can be encircled without encountering difficult regions. The total time is comparable, even if the navigation is smoother, because manual insertion / retraction was found more intuitive. Improvements in robotic insertion will be considered in the future.

Fig. \ref{subfig:target_right} illustrates the harder circumnavigation of the right pulmonary arteries. As shown by the timing color coding, there are no easy point-to-point transitions using the manual catheter. In contrast, robotic navigation around the first nine points can be performed quickly and easily. The second nine targets are more difficult. This arises because the current teleoperation scheme maps the operator's inputs into world frame motion of the magnet (see Section \ref{sec:actuation}). While this works well in most of the workspace, the mapping becomes nonintuitive for the last nine points of the right pulmonary veins. This problem can be eliminated in the future by implementing a teleoperation mapping that is based on the location of the catheter in the workspace.

Supplementary Video S1 compares manual and robotic navigation of the left and right pulmonary arteries in the phantom model. The challenges of navigation are particularly clear for the right pulmonary arteries of Task II. In point-to-point motion, it can be observed that, while the manual catheter tip does make contact somewhere within each target circle, the accuracy with which it reaches the center of the circle is considerably less than what is achieved with the robotic catheter. Furthermore, one can observe that the hand motions required to manipulate the manual catheter are more complex than those required for the robotic catheter.

\subsection{Soft Tissue Ablation}
\label{sub:ablation}
We performed ablation tests with a ball chain composed of 16 balls and compared the resultant lesions with the ones of a standard ablation catheter. The goal was to show that, without additional wires and electronics at the tip, a ball chain can achieve lesions comparable to the state of the art, with comparable power and duration \cite{Verma2021, Kumar2017} and compatible with a standard catheters used in the clinical practice (Tactiflex D-F, Abbott; 8FR diameter). 

Given the conductive nature of the magnetic balls' coating (Nickel plating), described at the beginning of Section \ref{sec:experiments}, we could simply connect the most proximal of the balls in the chain to an electrosurgical generator (Force 2, Valleylab). This avoids further cabling along the magnetic catheter, maximizing the magnetic volume - hence, force and torque - and minimizing the overall stiffness. 

The ball chain was covered with the silicone sleeve (details at the beginning of Section \ref{sec:experiments}) to test its insulation and heat resistance. Since we expect maximum heat at the interface of the chain tip and tissue, we used a shrink wrap to interface the sleeve and most distal ball (chain's tip). Bond between shrink wrap and ball was due to heat-induced shrinking and between the sleeve and the shrink wrap using glue. The shrink wrap created an insulation to guarantee bonding even at higher temperatures.

In the case of the standard manual catheter, we connected the tip electrode, via its native connection plug, to the electrosurgical generator. Since the ball chain does not allow for irrigation, we performed all our experiments without irrigation. This feature has been reported to affect depth and geometry of lesions \cite{Kumar2017}. From previous studies, since in absence of flow, we only considered that an effective lesion has to be transmural hence, be deeper than 2.9mm, in the thicker areas of the atrial wall; and have a width between 5.2mm and 6mm. In section \ref{sub:robotic_vs_manual}, we show that we can move between targets 5-6mm apart and create close paths around the pulmonary veins.  
\begin{figure}[t]
\includegraphics[width=\columnwidth, bb=0 0 200 150]{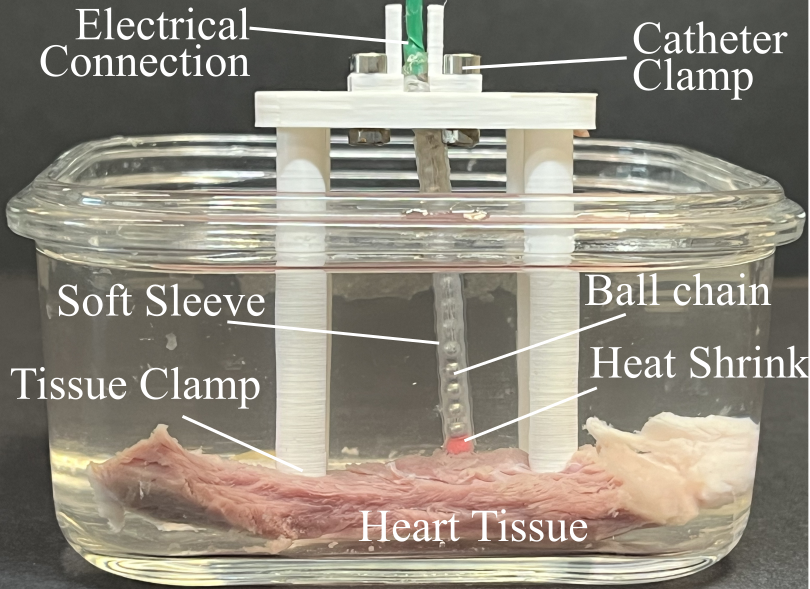}
\caption{Description of the experimental setup for cardiac ablation test.}
\label{fig:ablation_setup}
\end{figure}
~
\begin{figure}[t]
\includegraphics[width=\columnwidth, bb=0 0 250 100]{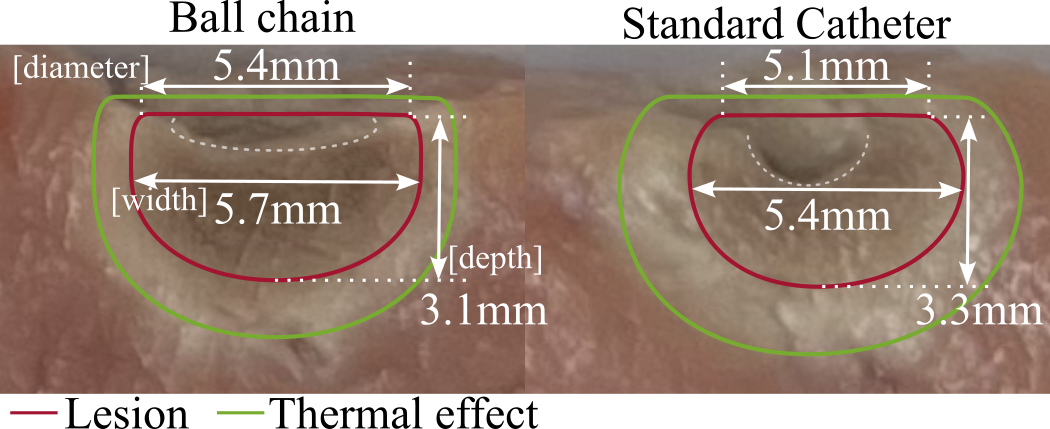}
\caption{Comparative analysis of ablation results with ball chain (left) and standard ablation catheter (right).}
\label{fig:ablation}
\end{figure}
The setup used for these experiments is represented in Fig. \ref{fig:ablation_setup}. A 3D printed clamp was glued on the rear end to a glass container to: secure the tissue to the bottom of the container (Tissue Clamp) and clamp the catheter/ball chain (Catheter Clamp) to guarantee constant contact with the tissue. We sliced tissue from a porcine heart to have a specimen of thickness larger than 1.5cm and have enough depth in the tissue to assess the full geometry of the lesions. For this reason we selected tissue from the ventricle, instead of the atrium where we would mostly perform ablation in the clinical practice, since thicker. The clamp guaranteed constant contact between the catheter/chain and the tissue; while we could not measure the applied force, we made sure contact was achieved. Specifically, in both cases, that the tip would create an indentation in the tissue of approximately 1mm depth, at the beginning of the trial. Tissue and catheter's/chain's tip were fully submerged in a solution of 0.9\% sodium chloride.

The goal of this experiment was to generate lesions of the required geometry \cite{Kumar2017}, in terms of diameter, width and depth (see Fig. \ref{fig:ablation} for definitions). The diameter is measured from the top of the lesion at the point of contact of catheter and tissue; the width is the largest area of the lesion in a cross-sectional view; and the depth is the level of penetration through the tissue, also measured in a cross-sectional view. The latter two measurements were taken by slicing the tissue in the center of the lesion, after ablation was performed. It should be noticed that the diameter and width do not match in general, as the lesion may expand diametrically as it penetrates the tissue \cite{Verma2021, Kumar2017}, when using a standard catheter. We expect this effect being related to the size and geometry of the catheter; relationship with irrigation flow has been reported in literature \cite{Kumar2017}, but we could not appreciate it in our experiments due to no irrigation.

To generate lesions comparable to the state-of-the art \cite{Kumar2017}, we ran the radiofrequency generator for the same time of 90s, while adjusting the power for the chain and standard catheter independently. For the ball chain we used 15W and for the standard catheter 30W; in both cases, we made sure that the power was compatible with what reported in literature and catheter's manual. 

Lesion assessment was performed visually, evaluating the area where tissue discoloration (dark brown zone) indicates full protein denaturation (red contour in Fig. \ref{fig:ablation}). This was also confirmed with tactile testing, as lesion presents as stiffer. The thermal effect encircled with the green line in Fig. \ref{fig:ablation} is a partial effect caused by lower temperature and does not indicates full lesion \cite{Verma2021}.

The results reported in Fig. \ref{fig:ablation} show a comparable lesion depth, which is also within the required range \cite{Kumar2017}. This is important to be able to create transmural lesions, and guarantee the overall depth of the tissue is insulated. It can be noticed that, using a standard catheter, the lesions tend to expand more diametrically but have a comparable width of 5.4mm with respect to the 5.7mm diameter of the ball chain. The lesions' depth was also beyond 2.9mm in both cases, indicating the ability of creating transmural lesion. We found comparable depth of 3.3mm for the standard catheter and 3.1mm for the ball chain.

We also notice a deeper indentation using the standard catheter, indicated by a fine dashed line in Fig. \ref{fig:ablation}, created by the ablating tip. This is related to the geometry of the catheter: while the chain tip is larger and smoother, the catheter has a small diameter and more pronounced edges. We made sure, in assessing the lesion that this was accounted for and measured the lesion from the top surface of the tissue. 

Further analysis of the effect of the tip geometry, size, irrigation and blood flow is required for in-depth understanding of type of lesion created. However, we can confidently conclude that the ball chain is able to create lesions comparable to the state-of-the-art catheters in terms of width and depth, by their natural ability of conducting currents on their surface. Therefore, they do not require further cabling which would make the system bulkier and introduce stiffening effects.

\section{Conclusion}

In this work, we investigated the usage of magnetic ball chains \cite{Pittiglio2023d} as ablation catheters for cardiac ablation and proposed a novel actuation method based on rotating permanent magnets. Thanks to the large magnetic content of the ball chain catheter, we showed that small-scale permanent magnets can be used for actuation. We proposed an actuation platform design where the actuating magnets can be placed on carts on each side of the patient. The carts are fixed in position during the operation but can be moved between procedures. This simplifies the integration with the clinical workflow, compared to using robotic manipulators to move the magnets \cite{Pittiglio2023a}, as it would have a smaller footprint and payload. The small scale of our target workspace, the heart, allows this platform to be effective, as we showed here.

In prior work, we showed how a ball chain's higher magnetic content enables bending tighter angles \cite{Pittiglio2023d}, when compared to designs using several discrete embedded magnets \cite{Edelmann2017, Jeon2018} or using catheters fabricated from a magnetic elastomeric matrix \cite{Kim2022, Pittiglio2022}. The present paper demonstrates how this property is useful for cardiac arrhythmia treatment which requires tight curvatures to reach locations such as the right pulmonary veins. We investigated the use of ball chains to generate clinically-relevant tip  forces without the need of large actuating magnets and showed that the chain is able to create ablation lesions comparable to clinical catheters  for cardiac ablation. Future work will investigate adding open irrigation, as in conventional ablation catheters.

To validate our proposed actuation concept, a small-scale prototype was presented and experiments were used to extrapolate the size of full-scale actuation units necessary for clinical use. Specifically, we used the force measurements from our experimental apparatus to calculate that a full-scale system could be made using two permanent magnets of 119mm diameter and 6.6kg mass.

In comparison experiments with a standard manual catheter on a common task in cardiac ablation (touching targets around the pulmonary veins), we observed that the robotic system can potentially improve the accuracy of lesion creation. We also noticed that the hand-eye coordination is not always intuitive, which suggests that better teleoperation strategies should be investigated in the future. To enable intelligent teleoperation, we will study integration of our catheter and actuation with real-time localization technologies \cite{Issa2019}. 

The paper shows high potential of the proposed platform to improve navigation in cardiac arrhythmia treatments. To further demonstrate this, we will work on building a full-scale system, integrate on-board irrigation, force sensing, and localization. These further developments will enable pre-clinical studies to demonstrate the full capabilities of the magnetic navigation method here introduced.

\ifCLASSOPTIONcaptionsoff
  \newpage
\fi

\bibliographystyle{ieeetr} 
\bibliography{literature.bib} 







\end{document}